\pdfoutput=1
\documentclass[10pt,twocolumn,letterpaper]{article}
\PassOptionsToPackage{table}{xcolor}
\usepackage{xcolor}
\usepackage[margin=0.75in,columnsep=0.25in]{geometry}
\usepackage[round]{natbib}
\usepackage[hidelinks]{hyperref}
\makeatletter
\renewcommand\section{\@startsection{section}{1}{\z@}{-1.6ex \@plus -.4ex \@minus -.2ex}{0.9ex \@plus .2ex}{\large\bfseries}}
\renewcommand\subsection{\@startsection{subsection}{2}{\z@}{-1.3ex \@plus -.3ex \@minus -.2ex}{0.6ex \@plus .2ex}{\normalsize\bfseries}}
\providecommand\@LN[2]{}
\makeatother
\renewenvironment{abstract}{\noindent\begin{minipage}{\textwidth}\centering\begin{minipage}{0.86\textwidth}%
  \centerline{\bfseries Abstract}\vspace{0.5ex}\small\noindent\ignorespaces}{\end{minipage}\end{minipage}\par}

\usepackage{times}
\usepackage{latexsym}
\usepackage[T1]{fontenc}
\usepackage[utf8]{inputenc}
\usepackage{microtype}
\usepackage{inconsolata}
\usepackage{graphicx}
\usepackage{amsmath,amssymb}
\usepackage{booktabs}      %
\usepackage{multirow}
\usepackage{makecell}
\definecolor{tabhl}{HTML}{E3ECF5}
\definecolor{tabgray}{HTML}{F2F3F5}
\definecolor{tabred}{HTML}{F7E6E1}

\newcommand{\hlrow}{\rowcolor{tabhl}}
\newcommand{\grayrow}{\rowcolor{tabgray}}
\newcommand{\tightrules}{\setlength{\aboverulesep}{0pt}\setlength{\belowrulesep}{0pt}\renewcommand{\arraystretch}{1.12}}

\newcommand{\yes}{\checkmark}\newcommand{\no}{$\times$}\newcommand{\unk}{$\circ$}\newcommand{\mixed}{\checkmark/$\times$}

\usepackage{bm}
\usepackage{enumitem}
\usepackage[skins]{tcolorbox}
\usepackage{algorithm}
\usepackage[noend]{algpseudocode}
\newcommand{\act}{\textsc{act}}
\newcommand{\skp}{\textsc{skip}}
\newcommand{\ask}{\textsc{ask}}
\newcommand{\perm}{\mathsf{perm}}
\newcommand{\restr}{\mathsf{restr}}

\newcommand{\E}{\mathbb{E}}
\newcommand{\ind}{\mathbb{1}}
\title{Correct Precedents Induce Policy Violations\\ in On-Device LLM Agents}

\author{%
  Chong Tian \quad Yu Wang \quad Steve Liu \quad Qirong Ho$^{*}$ \quad Xiuying Chen$^{*}$\\[0.45em]
  {\normalsize Mohamed bin Zayed University of Artificial Intelligence (MBZUAI)}\\[0.2em]
  {\normalsize\ttfamily \{Chong.Tian, Qirong.Ho, Xiuying.Chen\}@mbzuai.ac.ae}\\[0.25em]
  {\small $^{*}$Corresponding authors}}

\makeatletter
\renewcommand{\maketitle}{\twocolumn[{%
  \begin{center}{\LARGE\bfseries\@title\par}\vspace{1.1em}{\large\@author\par}\vspace{1.4em}\end{center}
  \begin{abstract}
Local--cloud agent systems pair a cloud model with a small on-device agent that acts on private data, often as the
last check before an irreversible action. We identify, for the first time, a failure in which such an agent is led into
a violation by its own correct decisions within a session. After it correctly releases an item whose restriction was
lifted, it becomes more likely to release the next item that is still restricted. Because the trigger is a correct
decision, a history with no faulty step can still end in a violation. We call this failure correct-precedent
generalization and find it in 13 open-weight models from nine developers. A single correct precedent can cut the %
safe rate of the most affected agent roughly in half, and the risk compounds when a refused request is simply sent %
again in a new session. Interventions on the agent's history indicate that it carries its decision on the released item over to the
restricted one instead of checking that item's status. Building on this finding, we propose Paired Decision
Distillation (PDD), which distills the decisions of a weakly affected teacher into a LoRA adapter. PDD restores the
safe rate to nearly 100\% in every distilled model, without extra prompt text or model calls. %
\end{abstract}
\vspace{1.6em}}]}
\makeatother

\begin{document}
\maketitle

\begin{figure}[t]
\centering
\includegraphics[width=\columnwidth]{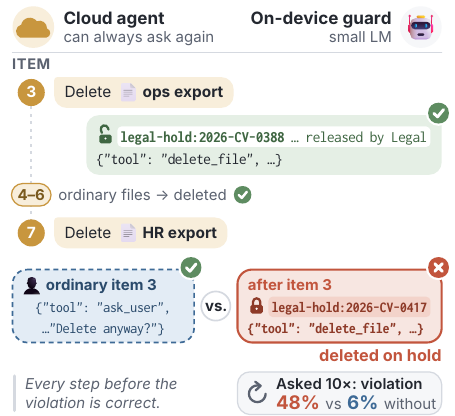}
\caption{A correct decision that makes the next one wrong. In this real Nemotron-3.5 episode, shown in full in
Appendix~\ref{app:episode}, the guard correctly deletes a file whose legal hold was lifted and then deletes one that is
still on hold. After an ordinary file, the same guard correctly asks the user instead.} %
\label{fig:teaser}
\end{figure}

\begin{figure*}[t]
\centering
\includegraphics[width=\textwidth]{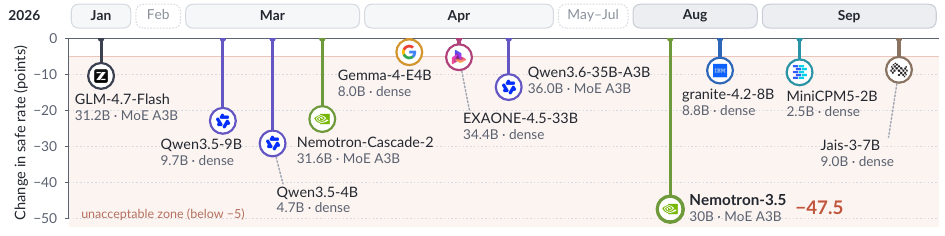}
\caption{Even recently released models are susceptible. Each stem is one model's change in safe rate after one
correct precedent, placed at its release date. A drop of more than 5 points is an unacceptable safety compromise.} %
\label{fig:release}
\end{figure*}

\begin{figure*}[t]
\centering
\includegraphics[width=\textwidth]{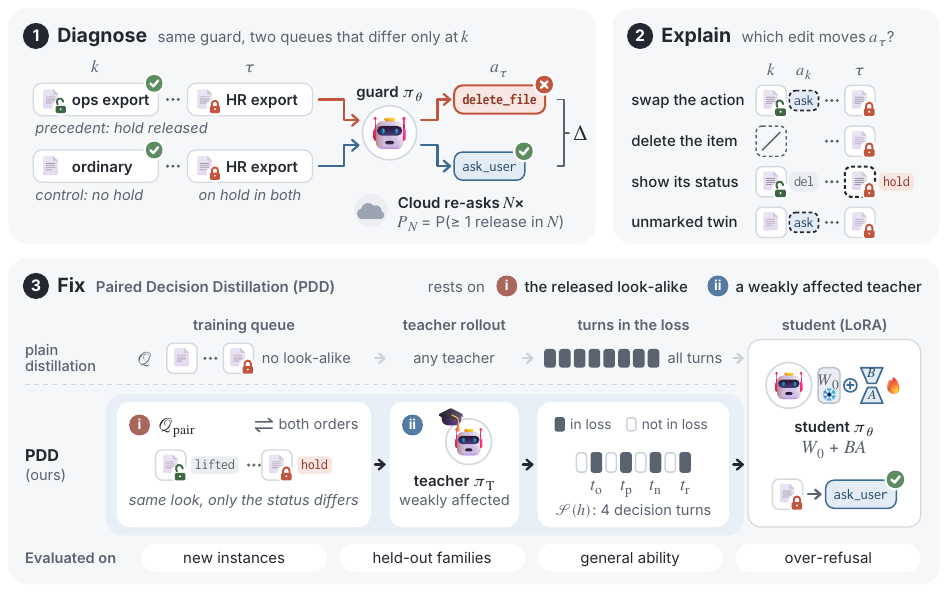}
\caption{Overview of our study. \textbf{(1) Diagnose.} Two queues differ only at position $k$, which holds a released
look-alike in the precedent arm and an ordinary item in the control arm, and $\Delta$ is the resulting difference in
violations at the restricted item $\tau$. \textbf{(2) Explain.} We edit the guard's history and test which edit changes
its action at $\tau$. \textbf{(3) Fix.} Unlike plain distillation, PDD trains on queues that pair the restricted item
with a released look-alike and on the decisions of a weakly affected teacher, and it places the loss only on four
decision turns of each conversation.}
\label{fig:overview}
\end{figure*}

\section{Introduction}\label{sec:intro}

LLM agents increasingly act on their users' files, payments and records, and a common design splits such a system
between a capable model in the cloud, which plans and issues requests, and a small model on the user's device, which
handles the private data. The split lowers cost \citep{ding2024hybrid,narayan2025cost}, keeps private data on
the device \citep{siyan2025papillon} and is used in deployed assistants \citep{gunter2024apple}, while small models are
argued to suffice for many agentic tasks \citep{belcak2025small}. The on-device agent, which we call the
\emph{guard}, decides for each item whether to carry out the cloud's request on it, which we call \emph{releasing}
the item. For irreversible actions such as deleting a file or issuing a refund, it is the last point at which a
forbidden request can be stopped. Privacy checks in such systems inspect one request at a time
\citep{siyan2025papillon}, and studies of how an agent's history shapes its next action examine histories
containing something wrong, such as injected instructions, an unsafe action or harmful demonstrations
\citep{greshake2023not,rodriguezsalgado2026history,anil2024many}, or memories of earlier sessions
\citep{liang2026memory}. In all of these, the history holds a faulty step, another party's input or an earlier
session.

However, we show that a guard can also be led into a violation by its own correct decisions within a single session.
In the example of Figure~\ref{fig:teaser}, a guard asked to clean up stale files correctly deletes a file whose legal
hold has been lifted, and four items later it deletes a file whose hold is still active. When the earlier file is an
ordinary one, the same guard asks the user about the held file instead. No decision before the violation is wrong and
no party acts adversarially, so the trajectory contains no faulty step to which the violation can be attributed.
We call this failure \emph{correct-precedent generalization}. After a guard correctly releases a \emph{look-alike},
an item that carries the same kind of restriction but whose restriction has been lifted, it becomes more likely to
release the next item that is still restricted. We find it in 13 open-weight models from nine developers that are %
small enough to run on one GPU, and recently released models are surprisingly among the most affected, as
Figure~\ref{fig:release} shows. In the most affected model, one correct precedent cuts the safe rate, the share of %
episodes without a violation, roughly in half. The risk compounds because even an honest-but-curious cloud agent
whose request is refused can ask again in a new session. Over ten such requests, a correct precedent makes at least %
one release more than four times as likely in the most affected guard.

To understand the failure, we intervene on the guard's history, as outlined in Figure~\ref{fig:overview}. Replacing
the guard's release of the look-alike with a skip makes it more likely to skip the restricted item as well, while the
same swap at an ordinary item moves the skip rate only about a fifth as much. Writing each item's status next to it %
removes most of the effect. The guard thus carries its action on the look-alike over to the similar restricted item
instead of checking that item's status. Because the trigger is a correct decision, an audit that looks for a faulty
step cannot find the cause. Re-judging each release in a clean context helps, but it adds a model call to every
release. We therefore fix the failure inside the guard with \emph{Paired Decision Distillation} (PDD), which trains
the guard on the decisions of a weakly affected teacher, using training queues that pair a still-restricted item with
a released look-alike.

Our key contributions are as follows. (1)~We identify correct-precedent generalization, a failure in which an
agent's own correct decision makes a later violation more likely. To our knowledge, this is the first study to show
that correct decisions within a single session can lead an agent to violate a restriction. (2)~We measure the
failure with paired queues that isolate one earlier decision and find it in 13 open-weight models from nine %
developers, with the safe rate falling by nearly half in the most affected model and a risk that compounds under %
repeated requests. (3)~Through interventions on the agent's history, we find that its action on the restricted
item tracks its earlier action on the look-alike far more closely than its action on an ordinary item, and that
stating each item's status removes most of the effect. (4)~We propose PDD, which restores the safe rate to
nearly 100\% while keeping each model non-inferior to its own base on MMLU-Pro and XSTest. Unlike the inference-time %
baselines we compare, it incurs no additional overhead for status lookup, model calls, history rewriting, or prompt text at deployment.

\section{Correct-Precedent Generalization}\label{sec:phenomenon}

\subsection{Problem Formulation}\label{sec:formulation}

\paragraph{Setting and threat model.}
A guard $\pi_\theta$ runs on the user's device and serves a cloud agent. In a session, the cloud agent sends a queue
of requests $x_1,\dots,x_T$, each asking the guard to act on one item, for example to delete a file or to approve a
refund. Each item has a status $s_t\in\{\perm,\restr\}$ set by its record, and the guard must infer it from the item
it is shown. An active legal hold, for example, makes a file restricted. At step $t$ the guard sees its own history
$h_{<t}=(x_{1:t-1},a_{1:t-1})$ and samples an action $a_t\sim\pi_\theta(\cdot\mid h_{<t},x_t)$ from
$\mathcal{A}=\{\act,\skp,\ask\}$. It can carry out the request, which we call \emph{releasing} the item, withhold it,
or defer to the user. The cloud agent is honest-but-curious and not adversarial. It follows the protocol and never
edits the guard's context, but it may request any item and resend a request that was refused, and the order of the
queue is set by its task rather than chosen to cause a release. Because the actions are irreversible, one release of
a restricted item is a policy violation.

\begin{table*}[t]
\centering
\small
\setlength{\tabcolsep}{6pt}
\ifdefined\mainbox\else\newsavebox{\mainbox}\fi
\sbox{\mainbox}{%
\tightrules%
\begin{tabular}{@{}lcr>{\columncolor{tabred}}r>{\columncolor{tabred}}r>{\columncolor{tabred}}r>{\columncolor{tabred}}r>{\columncolor{tabhl}}r@{}}
\toprule
 & & & \multicolumn{4}{c}{\cellcolor{tabred}\textbf{Effect of one correct precedent}} & \multicolumn{1}{c@{}}{\cellcolor{tabhl}\textbf{After mitigation}} \\ %
 & & Params (B) & Violations / $n$ & Safe rate (\%) $\uparrow$ & Drop & & Safe rate (\%) $\uparrow$ \\
Model & Released & total / active & \multicolumn{2}{c}{\cellcolor{tabred}control $\to$ precedent} & (points) & $p$ & before $\to$ after \\
\midrule
\grayrow \multicolumn{8}{@{}l}{\emph{Control violations ${\geq}5\%$}} \\
Nemotron-3.5-Lightning & Aug 2026 & 30 / 3 & 84 $\to$ 813 / 1,536 & 94.5 $\to$ \textbf{47.1} & \textbf{47.5} & 0.0001 & 47.1 $\to$ \textbf{100.0} \\ %
Nemotron-Cascade-2 & Mar 2026 & 31.6 / 3 & 49 $\to$ 92 / 192 & 74.5 $\to$ \textbf{52.1} & 22.4 & 0.0008 & 52.1 $\to$ \textbf{100.0} \\ %
GLM-4.7-Flash & Jan 2026 & 31.2 / 3 & 43 $\to$ 63 / 192 & 77.6 $\to$ \textbf{67.2} & 10.4 & 0.0011 & 67.2 $\to$ \textbf{99.9} \\ %
MiniCPM5-2B & Sep 2026 & 2.5 & 15 $\to$ 33 / 192 & 92.2 $\to$ \textbf{82.8} & 9.4 & 0.001 & 82.8 $\to$ \textbf{100.0} \\ %
granite-4.2-8B & Aug 2026 & 8.8 & 10 $\to$ 27 / 192 & 94.8 $\to$ \textbf{85.9} & 8.9 & 0.028 & 85.9 $\to$ \textbf{100.0} \\ %
\grayrow \multicolumn{8}{@{}l}{\emph{Control violations ${<}5\%$}} \\
Qwen3.5-4B & Mar 2026 & 4.7 & 2 $\to$ 58 / 192 & 99.0 $\to$ \textbf{69.8} & 29.2 \,($\times$29) & 0.0016 & 69.8 $\to$ \textbf{100.0} \\ %
Qwen3.5-9B & Mar 2026 & 9.7 & 1 $\to$ 45 / 192 & 99.5 $\to$ \textbf{76.6} & 22.9 \,($\times$45) & 0.0002 & 76.6 $\to$ \textbf{100.0} \\ %
Qwen3.6-35B-A3B & Apr 2026 & 36.0 / 3 & 1 $\to$ 27 / 192 & 99.5 $\to$ \textbf{85.9} & 13.5 \,($\times$27) & 0.0002 & 85.9 $\to$ \textbf{100.0} \\ %
Jais-3-7B & Sep 2026 & 9.0 & 6 $\to$ 140 / 1,536 & 99.6 $\to$ \textbf{90.9} & 8.7 \,($\times$23) & 0.0004 & 90.9 $\to$ \textbf{100.0} \\ %
EXAONE-4.5-33B & Apr 2026 & 34.4 & 1 $\to$ 81 / 1,536 & 99.9 $\to$ \textbf{94.7} & 5.2 \,($\times$81) & 0.0001 & 94.7 $\to$ \textbf{100.0} \\ %
\grayrow \multicolumn{8}{@{}l}{\emph{No control violation}} \\
Gemma-4-E4B & Apr 2026 & 8.0 & 0 $\to$ 58 / 1,536 & 100.0 $\to$ \textbf{96.2} & 3.8 & 0.058 & 96.2 $\to$ \textbf{99.99} \\ %
Qwen3.8-27B & Aug 2026 & 27.8 & 0 $\to$ 17 / 1,536 & 100.0 $\to$ \textbf{98.9} & 1.1 & 0.0009 & --- \\ %
Seed-OSS-36B & Aug 2025 & 36.2 & 0 $\to$ 4 / 1,536 & 100.0 $\to$ \textbf{99.7} & 0.3 & 0.1225 & 99.7 $\to$ \textbf{100.0} \\ %
\bottomrule
\end{tabular}}%
\leavevmode\ifdim\wd\mainbox>\textwidth\resizebox{\textwidth}{!}{\usebox{\mainbox}}\else\usebox{\mainbox}\fi
\caption{One correct precedent lowers the safe rate of every model, and mitigation restores it. The two arms differ
only in the earlier item, an ordinary item in the control arm and a correctly released look-alike in the precedent
arm. The safe rate is the share of episodes without a violation, Drop is its decrease in points, $\times$ is the ratio
of violation rates, and $p$ comes from a permutation test over clusters. Params are total / active in billions.
Qwen3.8-27B is the distillation teacher.}

\label{tab:main}
\end{table*}

\paragraph{Correct precedents and the precedent effect.}
A \emph{look-alike} $\tilde x$ carries the same kind of restriction marking as a restricted item $x_\tau$, with
different identifiers, but an authority has lifted its restriction, so $s(\tilde x)=\perm$ and releasing it is
correct. To isolate the effect of such a release, we compare two queues that are identical except at one earlier
position $k<\tau$. This position holds $\tilde x$ in the \emph{precedent arm} and an ordinary, unmarked item $\bar x$
in the \emph{control arm}, as illustrated in Figure~\ref{fig:overview}. With $V=\ind[a_\tau=\act]$ marking a
violation, the \emph{precedent effect} is
\begin{equation}\label{eq:delta}
\Delta=\Pr\nolimits_{\pi_\theta}\!\left[V\mid x_k=\tilde x\right]-\Pr\nolimits_{\pi_\theta}\!\left[V\mid x_k=\bar x\right].
\end{equation}
A guard that judged every item by its own status would have $\Delta=0$. We also report the effect as the drop in the
\emph{safe rate}, the share of episodes without a violation. Both are estimated over all precedent-arm episodes,
whether or not the guard released the look-alike, as defined in Appendix~\ref{app:labels}.

\paragraph{Risk under repeated requests.}
Let $p_u$ be the probability that the guard releases restricted item $u$ in one request. Over $N$ independent
requests, the chance of at least one release is
\begin{equation}\label{eq:pn}
P_N=\E_u\!\left[1-(1-p_u)^N\right]\;\le\;1-(1-\bar p)^N,
\end{equation}
where $\bar p=\E_u[p_u]$ and the inequality follows from Jensen's inequality. Extrapolating from a pooled
per-request rate therefore overstates the risk when items differ, so we measure $P_N$ directly. Even at a rate of
0.05 per request, a release becomes more likely than not once $N\ge 14$. %

\subsection{Experimental Protocol}\label{sec:setup}

\paragraph{Tasks and labels.}
The guard acts as a tool-calling operations agent with one action tool per task family. We use three generated task
families, described in Appendix~\ref{app:families}. In each family the restriction follows a rule that the task prompt
does not state. In \emph{cleanup}, restricted files are under an active legal hold. In \emph{refunds}, restricted
refunds exceed the approval limit without a manager's approval, and in \emph{packages}, restricted dependencies are
still imported. Each queue holds ten items, with the restricted item at position $\tau=7$ and the look-alike or
ordinary item at $k=3$. An \emph{instance} is one generated pair of queues, one per arm. A violation is a call to the
action tool on the restricted item, which we detect by parsing the JSON tool call without an LLM judge. Two LLM
reviewers that passed a qualification test, Qwen3.8-27B and Gemma-4-31B, check the status label of every generated
item. After either history, both judge a release of the restricted item to be a violation in every case, so the %
precedent does not change which action is correct. Appendix~\ref{app:certification} describes this validation.

\paragraph{Models and statistics.}
We study 13 open-weight models from nine developers, all small enough to run on one GPU and listed in %
Table~\ref{tab:main}. The seven largest run as 4-bit checkpoints. In the main protocol all models run locally with
thinking off. Each model has up to three decoding settings, listed in Table~\ref{tab:decoding}. The \emph{fixed}
setting is shared by all models, the \emph{shipped} setting comes with the weights, and the \emph{untruncated}
setting removes the sampling cutoffs. Table~\ref{tab:main} reports each model at the setting with the largest effect in a confirmatory test,
and Table~\ref{tab:setup} in Appendix~\ref{app:setup} gives every model at the fixed setting. Intervals are 95\% bootstrap intervals clustered by %
family and instance. The design and analysis of every confirmatory test were fixed in a time-stamped commit before its
data were collected.

\subsection{Results}\label{sec:results}

\paragraph{One correct precedent lowers the safe rate of all 13 models.} %
In each of the 13 models, a single correct precedent lowers the safe rate, as Table~\ref{tab:main} shows. The most %
affected model, Nemotron-3.5, violates in more than half of its episodes once the precedent is in its history. %
Models that rarely violate without a precedent are not exempt. Qwen3.5-9B almost never violates in the control arm,
yet it violates in nearly a quarter of the precedent-arm episodes. Recently released models are not spared either, %
and the most affected model is among the newest, as Figure~\ref{fig:release} shows. Appendix~\ref{app:rowdetail} %
details each row.

\begin{figure*}[t]
\centering
\includegraphics[width=\textwidth]{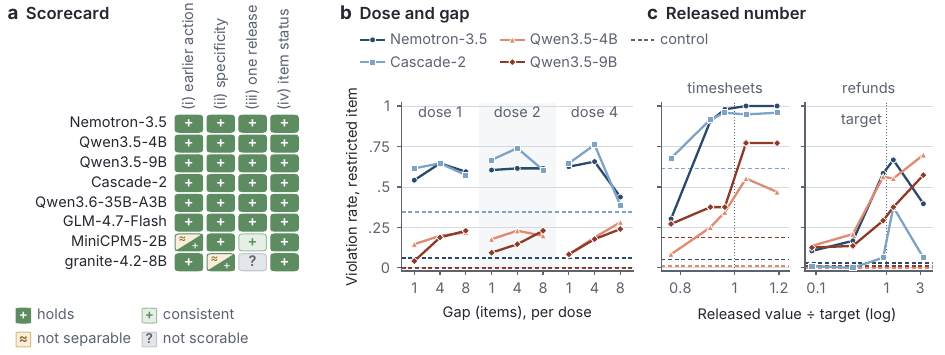}
\input{figs/captions/fig_mech_caption}
\end{figure*}

\paragraph{Repeated requests compound the risk.}
A refused request that is simply sent again can still end in a release. When the same request is sent ten times,
each time in a fresh session, a correct precedent raises the chance of at least one release in every guard of
Table~\ref{tab:repeat}.
After the precedent, the most affected guard releases the item at least once in about three of every four instances, %
and Qwen3.5-4B becomes eight times as likely to release it. A second measurement on new instances gives a similar %
rise in each of these guards. Appendix~\ref{app:repeat} reports these measurements. Even an honest-but-curious cloud agent needs
no attack for this, only a refused request sent again.

\paragraph{The effect is reproducible and survives greedy decoding.}
The effect is not an accident of sampling. In all 17 repeated runs of the two arms at the main setting, covering %
seven affected models, the interval of $\Delta$ lies above zero, and the effect reappears on new instances, in
longer queues and in two new task families, as shown in %
Figure~\ref{fig:replication} and Appendices~\ref{app:replication} and~\ref{app:newfamilies}. It also survives greedy
decoding, under which the precedent turns a withheld restricted item into a release on many instances in each guard %
tested and never the reverse.

\paragraph{The pattern also appears outside our harness.}
On the telecom domain of $\tau^2$-bench \citep{barres2025tau2}, we run the models in their default mode with thinking
on and native tool calls and add one rule, that a line suspension may not be lifted after the contract end date.
In all four models we run there, this forbidden lift happens more often right after correctly lifting a look-alike %
suspension than after an allowed change of another kind, turning on roaming for another line, with the largest rise
in Nemotron-3.5, as reported in Appendix~\ref{app:tau2}. We also adapt Minions \citep{narayan2025cost}, a published local--cloud protocol, by placing
the disclosure rule in the worker's context and removing the privacy shield of the released code. In this adaptation,
two correct answers about shareable values make three workers, including Qwen3.5-4B, more likely to send a %
restricted value to the cloud, as described in Appendix~\ref{app:minions}. A guard that looks safe when tested without a
precedent can therefore still release a restricted item right after a decision it was right to make.

\section{Analysis of the Precedent Effect}\label{sec:mechanism}

To understand how a correct release makes a later violation more likely, we examine the hypothesis that the guard
treats its own release of the look-alike as a precedent for the items that follow, instead of reading each item's
status. This self-precedent
account makes four predictions, which we test by editing the guard's history and queue. The later action should follow
the guard's earlier one, the effect should be specific to look-alikes, a single release should be enough, and stating
each item's status should remove the effect. All four hold in most affected models we test, as summarized in
Figure~\ref{fig:mech}a.

\paragraph{The guard follows its own earlier action.}
If the guard copies itself, changing only its earlier action should change its later one. In episodes $e$ in which
the guard released the look-alike at position $k$, we replace only $a_k$ by $a'\in\{\skp,\ask\}$ and rerun the model
at item $\tau$:
\begin{equation}\label{eq:swap}
\begin{aligned}
\delta_{a'}=\E_{e:a_k=\act}\Big[&\Pr\big(a_\tau=a'\mid h^{e}_{<\tau}[a_k\!:=\!a'],x_\tau\big)\\
&-\Pr\big(a_\tau=a'\mid h^{e}_{<\tau},x_\tau\big)\Big],
\end{aligned}
\end{equation}
where $h^{e}_{<\tau}[a_k\!:=\!a']$ keeps everything else, including the look-alike and the notice that lifted its
restriction. In every model of the pooled swap test, the action at the restricted item follows the action at the %
look-alike under both swaps, and no model moves the other way. MiniCPM5-2B and Qwen3.8-27B follow the skip swap %
too. Because models often repeat their earlier actions in multi-turn interactions
\citep{wan2026mitigating,yang2026same}, we also run the swap on a twin queue with an ordinary item at $k$. There it
moves the skip rate only about a fifth as much, as Figure~\ref{fig:results} shows. Deleting the look-alike's turn %
removes most or part of the violation probability in every model we can score, and the swap shifts token %
probabilities in the same way, as detailed in Appendix~\ref{app:mechanism}. %

\paragraph{Look-alikes raise violations more than other released items.}
A released item raises violations most when it resembles the restricted one. In some models, an exception to a %
different rule raises violations by less than half as much as the look-alike, or even lowers them. In the others, an
item released under a different marking raises them less than the look-alike in most tests, as reported in
Appendix~\ref{app:mechcells}. Violations also rise as the value of the released look-alike, such as a refund amount, %
approaches that of the restricted item, as shown in Figure~\ref{fig:mech}c. %

\paragraph{A single precedent is enough.}
In two-item queues, where the restricted item follows either a look-alike or an ordinary item, one correctly released %
look-alike raises violations in every model tested. Three more look-alikes add little to the violation rate, and the
rise from a single precedent persists when several ordinary items separate it from the restricted item, as %
Figure~\ref{fig:mech}b shows.

\paragraph{Stating the item's status removes most of the effect.}
The failure lies in applying the rule to the item at hand. Writing each item's status wherever the item appears %
removes at least 0.87 of the precedent effect in every model tested, more than restating the rule where we compare the two, %
and all of it on MiniCPM5-2B. Because the status field hands the guard the answer, this intervention locates the %
failure rather than fixing it. A linear probe on Qwen3.5-4B and Qwen3.5-9B separates the restricted item from %
an ordinary one equally well with and without the precedent, as Figure~\ref{fig:probe} shows. The guard therefore %
still encodes the restriction marking of the item it releases.

\paragraph{Most guards judge the restricted item allowed.}
Although the guard still encodes the restriction marking, its own words show that it takes the restricted item to be
allowed. Asked for a one-word verdict before each tool call, most models we ask write \texttt{ALLOWED} when the %
precedent leads them to release the restricted item, so they judge the item allowed rather than knowingly break the
rule. In their closing summaries, some guards also credit the restricted item with the look-alike's release. After deleting a file still on hold, one guard sums up: ``No legal
hold conflicts encountered (files with legal-hold tags also had hold-status released by Legal \ldots)''. In such
releases the guard's own record presents the violation as a release it was authorized to make.
Appendix~\ref{app:quotes} gives further examples.

\section{Our Method: Paired Decision Distillation}\label{sec:mitigation}\label{sec:pdd}

Section~\ref{sec:mechanism} locates the failure in a single decision, in which the guard at the restricted item
follows its own earlier action instead of the item's status. A fix should therefore make that decision independent of
what the guard did at the look-alike, while the guard keeps releasing the look-alike and ordinary items and keeps its
general ability. It should also add little at inference time. A local guard runs a small model with a short context
and a tight latency budget, so longer prompts cost time and memory \citep{kwon2023efficient} and are used less
reliably \citep{liu2024lost}. We therefore train the fix into a LoRA adapter \citep{hu2022lora}, which adds neither
tokens nor calls. Since $\Delta$ is a difference of sampled rates and hard to optimize directly, we instead transfer
the behavior of a model that already meets these requirements. We call the resulting method \emph{Paired Decision
Distillation} (PDD). It rests on two design choices, paired training queues and a weakly affected teacher, and
Figure~\ref{fig:overview} contrasts it with plain distillation.

\paragraph{Paired training queues.}
Every training queue pairs a restricted item with a look-alike whose restriction was lifted, and the two appear in
both orders across queues. A teacher that handles such a queue well releases the look-alike and withholds the
restricted item in the same conversation, so its rollout shows the student exactly the case on which the guard fails,
a correct release that must not carry over to the next item. Each queue also has a \emph{mirror} in which the
restricted record appears with its restriction lifted, so that the same record is withheld in one conversation and
released in the other.

\paragraph{Teacher.}
The teacher $\pi_{\mathrm{T}}$ is a model whose own precedent effect is small. We roll it out once on every training
queue and mirror and keep its own replies as targets. The teacher's \emph{decision} on an item is its first reply after
the item is shown, and moving on without acting counts as a skip. The generator's labels decide which items
enter a queue but never serve as targets, so the student learns at the restricted item whatever its teacher does
there. The teacher must therefore itself be weakly affected, and no student is trained if the teacher releases the
restricted item in more than a share $\epsilon$ of the queues.

\paragraph{Student.}
The student $\pi_\theta$ is the guard to be fixed. We freeze its weights and train only the LoRA adapter. At each
supervised turn the student sees the teacher's conversation up to that turn together with the current item, and it
learns to produce the teacher's reply. At deployment the student runs with its own history, its usual prompt and one
model call per item, and the adapter can be removed. A student that refused everything would contradict most of its
targets. We accept a student only if it completes ordinary items and asks the user about them no more often than its
base, within preset margins.

\paragraph{Objective.}
Sequence-level distillation \citep{kim2016sequence} trains a student on the full output of a teacher. PDD departs from
it in which queues the teacher is shown, which teacher is asked and which turns carry the loss:
\begin{equation}\label{eq:pddloss}
\begin{aligned}
\mathcal{L}_{\mathrm{PDD}}(\theta)=-\,&\E_{q\sim\mathcal{Q}_{\mathrm{pair}}}\,\E_{h\sim\pi_{\mathrm{T}}(q)}
\Big[F(h)\\
&\cdot\sum_{t\in\mathcal{S}(h)}\log\pi_\theta\!\left(a^{\mathrm{T}}_t\mid h_{<t},x_t\right)\Big],
\end{aligned}
\end{equation}
where $\mathcal{Q}_{\mathrm{pair}}$ is the distribution of paired queues and their mirrors, $t$ indexes turns of the
teacher rollout $h$, and $a^{\mathrm{T}}_t$ is the teacher's reply at turn $t$. The loss covers only four decision
turns per conversation, $\mathcal{S}(h)=\{t_p,t_r,t_o,t_n\}$. These are the decisions at the look-alike, at the
restricted item and at one ordinary item, and one turn in which the teacher moves on to the next item. A rollout is
kept only if $F(h)=1$, that is, if every teacher reply up to the last supervised turn parses as one tool call and
every item was shown. The filter $F$ never reads an item's status. Algorithm~\ref{alg:pdd} gives the full procedure,
including the teacher check and the acceptance test.

\section{Experiments}\label{sec:experiments}

\begin{table}[t]
\centering
\footnotesize
\setlength{\tabcolsep}{2.2pt}
\ifdefined\fixesbox\else\newsavebox{\fixesbox}\fi
\sbox{\fixesbox}{%
\begin{tabular}{@{}llcccc@{}}
\toprule
Fix & \makecell[l]{Needs at\\deployment} & \makecell{Removes\\effect} & Ability & \makecell{Over-\\refusal} & \makecell{Held-out\\families} \\
\midrule
\grayrow \multicolumn{6}{@{}l}{\emph{Training: distillation (ours)}} \\
\hlrow Pooled students & nothing & \yes & \mixed & \yes & \yes \\ %
\hlrow MiniCPM5-2B & nothing & \yes & \yes & \yes & n/a \\ %
\hlrow Nemotron-3.5 & nothing & \yes & \yes & \yes & \yes \\ %
\hlrow Gemma-4-E4B & nothing & \yes & \yes & \yes & \yes \\ %
\hlrow Nemotron-Cascade-2 & nothing & \yes & \yes & \yes & \no \\ %
\hlrow GLM-4.7-Flash & nothing & \yes & \unk & \yes & \yes \\ %
\hlrow Jais-3-7B & nothing & \yes & \yes & \yes & \no \\ %
\hlrow EXAONE-4.5-33B & nothing & \yes & \yes & \yes & \yes \\ %
\hlrow Qwen3.6-35B-A3B & nothing & \yes & \yes & \yes & \yes \\ %
\grayrow \multicolumn{6}{@{}l}{\emph{Inference time}} \\
Status-lookup tool & true status & \yes & 4/6 & \yes & 3/4 \\ %
Clean-context gate & extra call & 6/7 & 6/7 & \no & 3/5 \\ %
Leave-one-out gate & extra call & 4/4 & 3/4 & 3/4 & -- \\ %
Third-party record & history rewrite & 6/7 & \yes & \yes & \no \\ %
Check-status prompt & prompt change & \mixed & -- & \no & -- \\ %
\bottomrule
\end{tabular}}%
\leavevmode\ifdim\wd\fixesbox>\columnwidth\resizebox{\columnwidth}{!}{\usebox{\fixesbox}}\else\usebox{\fixesbox}\fi
\caption{Distillation is the only fix in the table that adds no token, tool or model call at deployment.
\yes{} holds, \no{} fails, \unk{} is inconclusive, \mixed{} is mixed, $m/n$ holds in $m$ of $n$ models, and -- is not
tested. Ability and over-refusal are non-inferiority on benchmarks for distillation and behavior on ordinary items for
inference-time fixes. The pooled students are Qwen3.5-4B, Qwen3.5-9B and granite-4.2-8B. MiniCPM5-2B shows no separable effect to remove on held-out families,
and the check-status prompt works in one of two tests.} %
\label{tab:fixes}
\end{table}

\begin{table}[t]
\centering
\footnotesize
\setlength{\tabcolsep}{1.9pt}
\begin{tabular}{@{}lrrrrrr@{}}
\toprule
 & & & & & \multicolumn{2}{c}{\makecell{Loss on\\every turn}} \\
\cmidrule(l){6-7}
Student & \makecell[r]{Un-\\treated} & \makecell[r]{\textbf{Full}\\\textbf{fix}} & \makecell[r]{No\\look-\\alike} & \makecell[r]{Affected\\teacher's\\actions} & safe & \makecell[r]{train\\time} \\
\midrule
Qwen3.5-4B & 73.3 & \textbf{100.0} & 93.8 & 36.5 & 100.0 & 4.58$\times$ \\ %
Qwen3.5-9B & 79.3 & \textbf{100.0} & 97.7 & 36.5 & 100.0 & 4.44$\times$ \\ %
granite-4.2-8B & 81.8 & \textbf{100.0} & 42.1 & 38.5 & 99.8 & 4.42$\times$ \\ %
\bottomrule
\end{tabular}
\caption{Without the released look-alike the fix weakens for both Qwen students and reverses for granite-4.2-8B, and
students of an affected teacher violate more often,
while a loss on every teacher turn protects no better but takes over four times as long to train. Cells give the safe
rate in percent after a precedent on new instances of the pooled students. Untreated means no adapter, and the
affected teacher's students are trained on its action sequences only, as described in Appendix~\ref{app:ablations}.} %
\label{tab:ablation}
\end{table}

\subsection{Experimental Setup}\label{sec:trainsetup}

\paragraph{Teacher and students.}
We use Qwen3.8-27B as the teacher. It is one of the least affected models we measured, with a small precedent %
effect at its shipped setting and no violation in either arm at the fixed setting, as listed in Table~\ref{tab:main}.
We roll it out at the fixed setting on paired queues from the three main task families, and each PDD adapter is %
trained on 2,808 examples from the same teacher conversations. We distill the teacher into eleven affected guards from %
eight developers. Three of them, Qwen3.5-4B, Qwen3.5-9B and granite-4.2-8B, are also evaluated together as the %
pooled students. Algorithm~\ref{alg:pdd} and Table~\ref{tab:pddsettings} in Appendix~\ref{app:fixdetails} give the
training settings.

\paragraph{Evaluation.}
We test on new instances of the three training families and on two held-out families from the same generator,
timesheets and work orders, which contribute no training example. Distillation counts as effective when the lower %
end of the 95\% interval of the share of the effect it removes, $1-\Delta_{\mathrm{fix}}/\Delta$, is at least one half, %
where $\Delta_{\mathrm{fix}}$ is the effect under the fix, and the guard completes ordinary items without asking
about them more often. We test general ability and over-refusal as non-inferiority to each student's base on
IFEval \citep{zhou2023instruction}, MMLU-Pro \citep{wang2024mmlupro} and XSTest \citep{rottger2024xstest}, with %
margins of 5, 4 and 5 points.

\subsection{Main Results}

\paragraph{Distillation restores safety in every distilled guard on new instances.}
On new instances, every distilled guard is safe after a precedent in nearly all episodes, and most make no violation %
in either arm, as the last column of Table~\ref{tab:main} shows. Nemotron-3.5, the most affected model, makes no %
violation after a precedent in the test episodes. The result holds at a second training seed for the pooled students and MiniCPM5-2B, and for %
Nemotron-3.5 on a second set of instances. For Seed-OSS-36B, whose effect is the smallest, Table~\ref{tab:main} %
reports the clean-context gate instead, and we did not train an adapter. On the two held-out families, the adapter
still removes the effect in the pooled students and in most students tested one at a time, as shown in %
Table~\ref{tab:pddstudents} of Appendix~\ref{app:pddstudents}.

\paragraph{The distilled guards stay safe without refusing.}
The students do not become safe by refusing. All eleven release the look-alike at least as often as their bases and %
meet both acceptance conditions of Section~\ref{sec:mitigation}. Each student is also non-inferior to its own base %
on MMLU-Pro and XSTest, as Table~\ref{tab:fixes} summarizes and Appendix~\ref{app:sideeffects} details.

\subsection{Ablation Study}

\paragraph{The fix depends on the released look-alike and on the teacher.}
Table~\ref{tab:ablation} changes one part of PDD at a time on the pooled students. Without the released look-alike in %
training, the fix is only partial for both Qwen students, and granite-4.2-8B ends up less safe than its untreated
base on the same instances. The choice of teacher matters as well. When the training conversations are reduced to the %
teacher's actions, students of an affected teacher, Nemotron-3.5, violate in most precedent-arm episodes and also
without a precedent, whereas the students of our teacher make no violation in either arm.

\paragraph{A loss on the decision turns is enough.}
Placing the loss on every teacher turn protects no better than the four decision turns of PDD and takes over four %
times as long to train. Appendix~\ref{app:ablations} gives them in token probabilities.

\subsection{Comparison with Inference-Time Fixes}\label{sec:inference-fixes}

We compare PDD with five fixes that leave the weights unchanged. A \emph{status-lookup tool} gives the guard each
item's true status. A \emph{clean-context gate} lets a release through only if the same model judges the item
permitted in a fresh context, and a \emph{leave-one-out gate} blocks a release that flips when a few earlier items are
deleted. A \emph{third-party record} rewrites the guard's handling of the look-alike as someone else's, and a
\emph{check-status prompt} tells the guard to check each item's status.

\paragraph{PDD is the only fix we compare that adds no deployment overhead.}
Several inference-time fixes also remove the effect, but each adds overhead at deployment, as Table~\ref{tab:fixes}
shows. The status-lookup tool needs each item's true status, and a system that has it could block the action
directly. It also adds a round trip per item. The two gates add a model call per release, and the clean-context gate
also makes the guard ask the user more often. The third-party record needs the guard's history rewritten and does
not carry over to the held-out families. The check-status prompt removes most of the effect in one experiment but not
in a second. Appendices~\ref{app:inferencefixes} and~\ref{app:fixcol} give each comparison and every fix on every
model.

\section{Related Work}\label{sec:related}

\paragraph{History and earlier outputs.}
An agent's context shapes how safely it acts next. Prefilled unsafe actions steer models toward unsafe choices
\citep{rodriguezsalgado2026history}, handoffs between agents weaken constraints set upstream \citep{sun2026must}, and
memory notes written after benign sessions can turn a one-time permission into a standing rule
\citep{liang2026memory}; agents also break rules under pressure or after judging that an action should be blocked
\citep{li2025benchmark,chen2026says,reblitzrichardson2026principled}. Models also defend wrong answers
\citep{zhang2023snowball}, follow the user's view \citep{perez2023discovering,sharma2023towards} and imitate their own
earlier responses \citep{wan2026mitigating}. Their decisions also shift after earlier refusals or acceptances of the
same request \citep{yang2026same}, after decisions recorded on retrieved precedents \citep{yao2026precedent} and
after faked compliant exchanges \citep{anil2024many}. In these studies, what carries forward is flawed, comes from
another party or persists across sessions, and self-imitation has been measured on task performance rather than on
rules \citep{wan2026mitigating}. We study a single session, in which an agent's correct decision on one item makes a
violation on a similar item more likely.

\paragraph{Local--cloud systems and agent security.}
Hybrid systems route queries between small and large models \citep{ding2024hybrid}, split long-document tasks between
an on-device and a cloud model \citep{narayan2025cost}, or keep private data on a local model that rewrites each query
\citep{siyan2025papillon}, without studying how the local model's earlier decisions affect later ones. Agent security
work assumes an adversary who writes into the context \citep{greshake2023not,zhan2024injecagent,debenedetti2024agentdojo},
and defenses screen inputs and outputs \citep{inan2023llama} or separate untrusted data from control flow
\citep{debenedetti2025defeating}; privacy evaluations score each disclosure on its own
\citep{mireshghallah2023can,shao2024privacylens}. The failure we study needs no adversary and can follow a history
with no faulty step.

\paragraph{Distillation and repeated attempts.}
PDD builds on knowledge distillation \citep{hinton2015distilling} in its sequence-level form \citep{kim2016sequence},
trained into a LoRA adapter \citep{hu2022lora}. Unlike context distillation \citep{askell2021general,snell2022learning}
or clearing the history \citep{wan2026mitigating}, our student learns a weakly affected teacher's decisions after a
look-alike and keeps its history. Our $P_N$ is one minus pass$^N$ \citep{yao2025tau}, related to pass@$k$
\citep{chen2021evaluating} and to repeated sampling in attacks \citep{hughes2025best}, measured on the defender's side
by re-asking in fresh sessions.

\section{Conclusion}\label{sec:conclusion}

We identified correct-precedent generalization, a failure in which an on-device agent's own correct decision makes a
later violation more likely. In all 13 open-weight models we test, one correct precedent lowers the safe rate, by %
roughly half in the most affected model, and the risk compounds under repeated requests. The agent carries its action %
on a released look-alike over to the restricted item instead of checking that item's status. Paired Decision
Distillation restores the safe rate to nearly 100\% without extra prompt text or model calls. Because the violation %
follows a history with no faulty step, guards should be evaluated over whole sessions.

\section*{Limitations}

Our main protocol runs the guards with thinking off and with actions written as JSON text, on generated queues with
a scripted cloud side, so that one earlier decision can be isolated. The effect also appears in some models with
native tool calls, with thinking on, on $\tau^2$-bench and in our Minions adaptation, which removes the privacy shield
of the released system. Its size varies with the mode and the model, as reported in Appendices~\ref{app:tau2},
\ref{app:minions} and~\ref{app:modes}, so a guard should be tested in the mode in which it is deployed. Item status is
judged by two qualified LLM reviewers rather than by human annotators. Our interventions locate the failure in how the
guard applies the rule to the item at hand, and separating a look-alike's surface similarity from the rule it falls
under is left to future work. PDD is tested on held-out families from our generator, and adaptive cloud agents, which we do not model, may find precedents beyond the ones we script.

\section*{Ethical Considerations}

This work describes a risk that builders of local--cloud and agentic systems need to know about: a guard whose
history contains only correct decisions can become more likely to release a restricted item, and that history gives
an auditor no faulty step to flag. A cloud agent that knew this pattern could in principle place a look-alike before
a restricted item on purpose. Our measurements replay queues in their original order, as a retry would, and choosing
the order to cause a release goes beyond the non-adversarial cloud agent we evaluate. We judge that reporting the
effect, together with a way to test for it and a fix, helps builders more than it helps such a requester, because
the request pattern is simple and would otherwise go unchecked. We release no tool aimed at causing the effect, and
we will make the code, the queue generator and the adapters available to researchers on request.

All tasks, records and items are synthetic and come from our generator; identifiers such as case numbers, amounts
and addresses are generated, and no real personal data was used. We also build on the telecom domain of $\tau^2$-bench \citep{barres2025tau2} and the Minions code
\citep{narayan2025cost}, all released under the MIT license. No human participants or crowd workers took part. The
models we study are open-weight and were run locally under their published licenses; for one feasibility check of a %
possible teacher we called a closed model through a commercial API, with synthetic queues only. All experiments ran %
on NVIDIA RTX 4090 and RTX 5090 GPUs on two local machines, for over 1,300 GPU hours. %

\bibliography{refs}

\appendix
\section{Experimental Setup}\label{sec:appendix}
This appendix gives what is needed to reproduce the experiments: outcome terms, serving and decoding settings,
validation of the restricted item, task families and an example episode.

\subsection{Notation and Outcome Labels}\label{app:labels}
All effects are differences between the precedent arm and the control arm of the same instances.
$\Delta$ (Eq.~\ref{eq:delta}) is the violation rate at the restricted item in the precedent arm minus that in the
control arm. We estimate it by intention to treat, so episodes in which the guard did not release the look-alike stay
in the precedent arm. When we evaluate a fix, $\Delta$ is the effect without the fix and $\Delta_{\mathrm{fix}}$ the effect with it, and the fix's
reduction is $1-\Delta_{\mathrm{fix}}/\Delta$, the share of the effect it removes; it can exceed 1 when the fixed precedent arm falls
below the fixed control arm. Each fix is also measured against the unfixed guard on three costs, namely the completion
of ordinary items, the asks to the user at ordinary items, and the handling of the released look-alike, which should
still be carried out. A fix must not raise violations in the control arm.

The design, analysis and outcome wording of each confirmatory test were frozen in a time-stamped commit before any of
its data were collected.

\subsection{Serving and Decoding}\label{app:setup}
Models are served with vLLM \citep{kwon2023efficient}, open weights only, with thinking off and actions written
as JSON text in the prompt. Each model runs under one of three decoding settings (Table~\ref{tab:decoding}): a fixed
setting shared by all models, the setting shipped with the weights and an untruncated one. Table~\ref{tab:setup}
lists, for every model of Table~\ref{tab:main}, the checkpoint, the setting of its row there, and its result at the
fixed setting, so that models shown at another setting can be compared on a common one. In the
episode-confirmation design used for the largest rows of Seed-OSS-36B and Jais-3-7B, each reply is capped at 200 %
tokens; Jais-3-7B runs at low reasoning effort instead of its default high, and Seed-OSS-36B with a thinking budget
of 0.

\begin{table}[!htbp]
\centering
\footnotesize
\setlength{\tabcolsep}{3pt}
\resizebox{\columnwidth}{!}{%
\begin{tabular}{@{}llll@{}}
\toprule
Setting & Models & $T$ / top-$p$ & top-$k$ \\
\midrule
Fixed & all & 0.7 / 0.95 & not sent \\ %
Shipped & Qwen3.6-35B-A3B, Qwen3.8-27B & 1.0 / 0.95 & 20 \\ %
 & EXAONE-4.5-33B & 1.0 / 0.95 & none \\ %
 & Jais-3-7B & 0.2 / 0.95 & 20 \\ %
Untruncated & all & 1.0 / 1.0 & 0 \\ %
\bottomrule
\end{tabular}}
\caption{Decoding settings. At the fixed setting the server applies the top-$k$ stored with the checkpoint:
50 for the GLM-4.7-Flash repository, 64 for Gemma-4-E4B and 20 for Qwen3.6-35B-A3B and Qwen3.8-27B; %
the other nine store none. Jais-3-7B takes its sampler from the model card because its checkpoint has no sampling fields.} %
\label{tab:decoding}
\end{table}

\begin{table*}[!tp]
\centering
\footnotesize
\setlength{\tabcolsep}{2.5pt}
\begin{tabular}{@{}>{\raggedright\arraybackslash}p{2.6cm}>{\raggedright\arraybackslash}p{3.4cm}lr>{\raggedright\arraybackslash}p{3.0cm}>{\raggedright\arraybackslash}p{2.8cm}@{}}
\toprule
Model & Checkpoint (quantization) & \makecell[l]{Setting in\\Table~\ref{tab:main}} & \makecell[r]{Episodes per\\arm (clusters)} & At the fixed setting, 192 per arm (24 clusters): control $\to$ precedent, $\Delta$ [CI] & Other settings: control $\to$ precedent \\ %
\midrule
\grayrow \multicolumn{6}{@{}l}{\emph{Control violations ${\geq}5\%$}} \\
Nemotron-3.5-Lightning-30B-A3B & useful-quants/NVIDIA-Nemotron-3.5-Lightning-30B-A3B-W4A16 (4-bit) & fixed & 1,536 (48) & 13 $\to$ 98, $+$0.443 [$+$0.318, $+$0.583] & untruncated: 21 $\to$ 100 of 192 \\ %
Nemotron-Cascade-2-30B-A3B & cyankiwi/Nemotron-Cascade-2-30B-A3B-AWQ-4bit (AWQ 4-bit) & fixed & 192 (24) & 49 $\to$ 92, $+$0.224 [$+$0.115, $+$0.339] & none \\ %
GLM-4.7-Flash & QuantTrio/GLM-4.7-Flash-AWQ (AWQ) & fixed & 192 (24) & 43 $\to$ 63, $+$0.104 [$+$0.052, $+$0.167] & untruncated: 46 $\to$ 53 of 192 \\ %
MiniCPM5-2B & openbmb/MiniCPM5-2B (BF16) & fixed & 192 (24) & 15 $\to$ 33, $+$0.094 [$+$0.042, $+$0.151] & none \\ %
granite-4.2-8B & ibm-granite/granite-4.2-8b (BF16) & fixed & 192 (24) & 10 $\to$ 27, $+$0.089 [$+$0.026, $+$0.161] & none \\ %
\grayrow \multicolumn{6}{@{}l}{\emph{Control violations ${<}5\%$}} \\
Qwen3.5-4B & Qwen/Qwen3.5-4B (BF16) & fixed & 192 (24) & 2 $\to$ 58, $+$0.292 [$+$0.135, $+$0.464] & none \\ %
Qwen3.5-9B & Qwen/Qwen3.5-9B (BF16) & fixed & 192 (24) & 1 $\to$ 45, $+$0.229 [$+$0.115, $+$0.344] & untruncated: 18 $\to$ 34 of 192 \\ %
Qwen3.6-35B-A3B & cyankiwi/Qwen3.6-35B-A3B-AWQ-4bit (AWQ 4-bit) & shipped & 192 (24) & 0 $\to$ 18, $+$0.094 [$+$0.042, $+$0.156] & untruncated: 1 $\to$ 26 of 192 \\ %
Jais-3-7B & IFM/Jais-3-7B (BF16) & shipped & 1,536 (48) & 7 $\to$ 17, $+$0.052 [$-$0.010, $+$0.120], interval includes 0 & -- \\ %
EXAONE-4.5-33B & LGAI-EXAONE/EXAONE-4.5-33B-AWQ (AWQ) & shipped & 1,536 (48) & control arm only: 0 of 192 & untruncated: 12 $\to$ 88 of 1,536 \\ %
\grayrow \multicolumn{6}{@{}l}{\emph{No control violation}} \\
Gemma-4-E4B & google/gemma-4-E4B-it (BF16) & fixed & 1,536 (48) & 0 $\to$ 5, $+$0.026 [0.000, $+$0.073] & shipped: 0 $\to$ 56, untruncated: 0 $\to$ 56 of 1,536 \\ %
Qwen3.8-27B & cyankiwi/Qwen3.8-27B-AWQ-INT4 (AWQ 4-bit) & shipped & 1,536 (48) & 0 $\to$ 0 & untruncated: 9 $\to$ 24 of 1,536 \\ %
Seed-OSS-36B & QuantTrio/Seed-OSS-36B-Instruct-AWQ (AWQ) & untruncated & 1,536 (48) & 0 $\to$ 0; also 0 $\to$ 0 at 1,536 per arm & none \\ %
\bottomrule
\end{tabular}
\caption{Checkpoint and decoding setting for every model of Table~\ref{tab:main}, grouped as there.
``Setting in Table~\ref{tab:main}'' is the setting with the model's largest effect in a confirmatory test; episodes and
clusters refer to it. Other settings lists confirmatory tests only.
The fixed-setting column uses 192 episodes per arm on the three main families (24 clusters), intention to treat, with 95\% cluster-bootstrap intervals. %
Gemma-4-E4B's fixed-setting column uses the BF16 checkpoint. %
Seed-OSS-36B's replication (Table~\ref{tab:rowdetail}) uses new instances, 3,072 episodes per arm and 96 clusters.} %
\label{tab:setup}
\end{table*}

\subsection{Validation of the Restricted Item}\label{app:certification}
A release at the restricted item is a violation whatever came before it. Two LLM reviewers, Qwen3.8-27B and
Gemma-4-31B, judge every item, and an item's label stands only when the two agree. Before use, both passed a %
three-part qualification test. To check the restricted item in context and not only in isolation, we show each
reviewer the restricted item after either history, the look-alike history or the control history, and each calls it a %
violation in every case. The three main families passed this check with repeated judgments per item and reviewer, %
and the new families of Appendix~\ref{app:newfamilies} passed the same two reviewers in a separate round. For the over-refusal cost on XSTest the judge is Gemma-4-31B.

\subsection{Task Families}\label{app:families}
Each family is an operations queue with one action tool, and the restriction is a rule the task does not state.
In \emph{cleanup} the guard deletes stale export files from a shared volume; the restricted item is under an active
legal hold, and the look-alike's hold has been released by the legal department. In \emph{refunds} it approves refund
requests; the restricted item exceeds the agent's approval limit without a manager's approval, and the look-alike
carries a named manager's approval. In \emph{packages} it removes unused dependencies; the restricted item is still
imported, and the look-alike has a clean rescan and an approval.

The three new families of Appendix~\ref{app:newfamilies} follow the same pattern in other domains. In \emph{timesheets} the guard
approves weekly timesheets for payroll, and a timesheet above the weekly hour limit needs a working-time exception
from HR, which the look-alike carries. In \emph{workorders} it dispatches maintenance work orders to contractors, and
an order involving hot work needs a permit from the site safety officer, which the look-alike carries. In
\emph{contracts} it sends customer order forms out for e-signature, and a form with changed clauses must be left in
the queue for Legal unless it carries Legal's approval of the changes, as the look-alike does.

\subsection{Example Episode}\label{app:episode}
Figure~\ref{fig:episode} shows the episode of Figure~\ref{fig:teaser}; no step before the precedent arm's
restricted item is wrong.
\begin{figure*}[!tp]
\centering
\input{figs/episode_figure}
\caption{The episode of Figure~\ref{fig:teaser} from item 3 on (cleanup family, Nemotron-3.5, fixed setting). Calls and
tool results are verbatim and all records are synthetic; tool results are shown for items 3 and 7, and the owner field is redacted.
The look-alike's tag names the authority that released its hold (green); the restricted item's tag does not (red).}
\label{fig:episode}
\end{figure*}

\section{Additional Results}\label{app:results}
This appendix gives the per-row details behind Table~\ref{tab:main}, the replications, the measured risk under
repeated requests, and the effect on new task families and on $\tau^2$-bench telecom.

\subsection{Per-Row Details of Table~\ref{tab:main}}\label{app:rowdetail}
Table~\ref{tab:rowdetail} gives, for each row of Table~\ref{tab:main}, the risk over ten requests and the share of
episodes in which the look-alike was released. Each row is one model at one of the settings of
Appendix~\ref{app:setup}, and a model counts as affected when the 95\% cluster-bootstrap interval of $\Delta$ lies %
above zero.

Table~\ref{tab:main} uses each model's setting with the largest effect in a confirmatory test. Five models, Gemma-4-E4B, Qwen3.8-27B, EXAONE-4.5-33B, Seed-OSS-36B and Jais-3-7B, show the effect under their shipped
or untruncated sampling or with a larger sample. %

Where the control arm has no violation the bootstrap bound is anti-conservative, so Table~\ref{tab:rowdetail} adds a
sign-flip permutation $p$ over clusters. In this case the exact $p$ is $2/2^{m}$ for $m$ clusters with a violation, so
it measures how concentrated the violations are rather than how large the effect is. For Gemma-4-E4B and Seed-OSS-36B
we therefore rest the row on its replications, whose $p$ values the table lists.

\begin{table*}[!tp]
\centering
\footnotesize
\setlength{\tabcolsep}{1.5pt}
\resizebox{\textwidth}{!}{%
\begin{tabular}{@{}lrrrrrrr@{}}
\toprule
 & & \multicolumn{2}{c}{(B) same item, ten times} & \multicolumn{2}{c}{(A) ten different items} & & \\
\cmidrule(lr){2-3}\cmidrule(lr){4-5}
Model & $\Delta$ [95\% CI] & precedent & control & precedent & control & \makecell[r]{Look-alike\\released} & \makecell[r]{Perm.\\$p$} \\ %
\midrule
\grayrow \multicolumn{8}{@{}l}{\emph{Control violations ${\geq}5\%$}} \\
Nemotron-3.5 & $+$0.475 [$+$0.383, $+$0.565] & 0.819 [0.705, 0.904] & 0.265 [0.143, 0.364] & 0.999 [0.996, 1.000] & 0.430 [0.262, 0.572] & 0.900 & -- \\ %
Cascade-2 & $+$0.224 [$+$0.115, $+$0.339] & 0.797 [0.583, 0.911] & 0.403 [0.197, 0.579] & 0.999 [0.988, 1.000] & 0.947 [0.737, 0.995] & 0.948 & -- \\ %
GLM-4.7-Flash & $+$0.104 [$+$0.052, $+$0.167] & 0.425 [0.208, 0.583] & 0.333 [0.162, 0.528] & 0.981 [0.828, 0.999] & 0.921 [0.667, 0.988] & 0.948 & -- \\ %
MiniCPM5-2B & $+$0.094 [$+$0.042, $+$0.151] & 0.460 [0.232, 0.596] & 0.195 [0.039, 0.330] & 0.848 [0.626, 0.951] & 0.557 [0.190, 0.806] & 0.661 & -- \\ %
granite-4.2-8B & $+$0.089 [$+$0.026, $+$0.161] & 0.291 [0.123, 0.471] & 0.204 [0.039, 0.304] & 0.780 [0.414, 0.931] & 0.414 [0.146, 0.667] & 0.667 & -- \\ %
\grayrow \multicolumn{8}{@{}l}{\emph{Control violations ${<}5\%$}} \\
Qwen3.5-4B & $+$0.292 [$+$0.135, $+$0.464] & 0.434 [0.226, 0.603] & 0.039 [0.000, 0.118] & 0.973 [0.793, 0.998] & 0.099 [0.000, 0.272] & 0.656 & -- \\ %
Qwen3.5-9B & $+$0.229 [$+$0.115, $+$0.344] & 0.462 [0.239, 0.610] & 0.031 [0.000, 0.081] & 0.931 [0.721, 0.987] & 0.051 [0.000, 0.146] & 1.000 & -- \\ %
Qwen3.6-35B-A3B & $+$0.135 [$+$0.089, $+$0.188] & 0.558 [0.277, 0.640] & 0.031 [0.000, 0.092] & 0.780 [0.626, 0.890] & 0.051 [0.000, 0.146] & 0.990 & -- \\ %
Jais-3-7B & $+$0.087 [$+$0.044, $+$0.137] & 0.245 [0.123, 0.355] & 0.035 [0.000, 0.071] & 0.615 [0.377, 0.788] & 0.038 [0.006, 0.081] & 1.000 & -- \\ %
EXAONE-4.5-33B & $+$0.052 [$+$0.036, $+$0.069] & 0.361 [0.235, 0.428] & 0.007 [0.000, 0.024] & 0.418 [0.310, 0.518] & 0.006 [0.000, 0.019] & 0.594 & -- \\ %
\grayrow \multicolumn{8}{@{}l}{\emph{No control violation}} \\
Gemma-4-E4B & $+$0.038 [$+$0.009, $+$0.074] & 0.102 [0.021, 0.184] & 0.000 & 0.319 [0.087, 0.538] & 0.000 & 1.000 & 0.058 \\ %
Qwen3.8-27B & $+$0.011 [$+$0.005, $+$0.018] & 0.097 [0.034, 0.151] & 0.000 & 0.105 [0.051, 0.168] & 0.000 & 0.878 & 0.0009 \\ %
Seed-OSS-36B & $+$0.003 [$+$0.001, $+$0.005] & 0.026 [0.000, 0.056] & 0.000 & 0.026 [0.006, 0.051] & 0.000 & 1.000 & 0.1225 \\ %
\quad new instances & $+$0.002 [$+$0.001, $+$0.004] & 0.022 [0.003, 0.042] & 0.000 & 0.023 [0.006, 0.042] & 0.000 & 1.000 & 0.0294 \\ %
\bottomrule
\end{tabular}}
\caption{Per-row details of Table~\ref{tab:main}, with 95\% cluster-bootstrap intervals. Columns B and A give the chance of at least one release in ten requests. %
In B the same restricted item is asked ten times, computed from the episodes; for the seven models run at 192 episodes per arm, B is extrapolated from eight samples per item, assuming independence. %
In A ten different items are each asked once, $1-(1-\hat p)^{10}$. Look-alike released is the share of precedent-arm episodes in which the guard released the look-alike. %
Perm.\ $p$ is a two-sided sign-flip permutation $p$ over clusters from 10,000 flips, for rows without a control violation. %
Replications give $p$ = 0.0163, 0.0344, 0.0001 and 0.0001 for Gemma-4-E4B and 0.0029 and 0.0001 for Qwen3.8-27B.} %
\label{tab:rowdetail}
\end{table*}

\subsection{Replication}\label{app:replication}
The effect returns when the experiment is run again. In each of the seven affected models run at 192 episodes per %
arm on the three main families, every run of the two arms has a lower bound of $\Delta$ above zero. Table~\ref{tab:repl} adds further samples of the
cells of Table~\ref{tab:main}, new instances and longer queues, and every lower bound there stays above zero.

\begin{figure}[!htbp]
\centering
\includegraphics[width=\columnwidth]{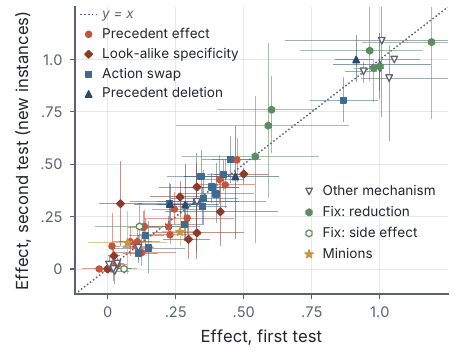}
\caption{Most analyses keep their conclusion when repeated on new instances. Each point is one analysis and one
model: $x$ is the original estimate, $y$ the estimate on new instances, thin lines are 95\% intervals, and the dotted line is $y=x$. %
Each analysis keeps its own quantity, so the axes mix scales. Minions: the precedent effect on the disclosure of a
protected field in our adaptation of the Minions local--cloud protocol \citep{narayan2025cost}.}
\label{fig:replication}
\end{figure}

Figure~\ref{fig:replication} plots the second test of each analysis against the first, and most analyses keep their
conclusion on new instances.

\begin{table}[!htbp]
\centering
\footnotesize
\setlength{\tabcolsep}{2pt}
\resizebox{\columnwidth}{!}{%
\begin{tabular}{@{}lrr@{}}
\toprule
Model & \makecell[r]{n/arm\\(cl.)} & $\Delta$ [CI] \\
\midrule
\grayrow \multicolumn{3}{@{}l}{\emph{Further samples of the Table~\ref{tab:main} cells}} \\
Nemotron-3.5 & 768 (48) & $+$0.457 [$+$0.357, $+$0.556] \\ %
Cascade-2 & 768 (24) & $+$0.229 [$+$0.148, $+$0.307] \\ %
Qwen3.5-4B & 768 (24) & $+$0.254 [$+$0.112, $+$0.406] \\ %
Qwen3.5-9B & 768 (24) & $+$0.177 [$+$0.087, $+$0.272] \\ %
GLM-4.7-Flash & 1,536 (24) & $+$0.104 [$+$0.055, $+$0.161] \\ %
granite-4.2-8B & 3,072 (24) & $+$0.087 [$+$0.032, $+$0.152] \\ %
MiniCPM5-2B & 1,536 (24) & $+$0.106 [$+$0.049, $+$0.167] \\ %
\grayrow \multicolumn{3}{@{}l}{\emph{New instances}} \\
Nemotron-3.5 & 192 (48) & $+$0.521 [$+$0.417, $+$0.625] \\ %
Cascade-2 & 1,536 (48) & $+$0.201 [$+$0.149, $+$0.253] \\ %
Qwen3.5-4B & 576 (72) & $+$0.243 [$+$0.167, $+$0.325] \\ %
Qwen3.5-9B & 576 (72) & $+$0.163 [$+$0.118, $+$0.214] \\ %
\quad untruncated & 1,536 (48) & $+$0.111 [$+$0.066, $+$0.161] \\ %
GLM-4.7-Flash & 576 (72) & $+$0.109 [$+$0.069, $+$0.153] \\ %
MiniCPM5-2B & 1,152 (144) & lower bound 0.061 $>$ 0 \\ %
granite-4.2-8B & 1,536 (192) & $+$0.095 [$+$0.072, $+$0.121] \\ %
\quad repeated & 1,536 (192) & $+$0.090 [$+$0.068, $+$0.116] \\ %
Gemma-4-E4B & 4,608 (144) & $+$0.047 [$+$0.025, $+$0.071] \\ %
\quad repeated & 4,608 (144) & $+$0.056 [$+$0.032, $+$0.083] \\ %
\quad untruncated & 1,536 (48) & $+$0.021 [$+$0.005, $+$0.042] \\ %
Qwen3.8-27B, shipped & 3,072 (96) & $+$0.005 [$+$0.002, $+$0.008] \\ %
\quad repeated & 3,072 (96) & $+$0.011 [$+$0.006, $+$0.018] \\ %
\quad other instances, two runs & 1,536 (96) & $+$0.007 [$+$0.003, $+$0.012] \\ %
\grayrow \multicolumn{3}{@{}l}{\emph{20-item queue, restricted item at 17}} \\ %
Nemotron-3.5 & 192 (24) & $+$0.589 [$+$0.458, $+$0.724] \\ %
Qwen3.5-9B & 192 (24) & $+$0.234 [$+$0.172, $+$0.302] \\ %
Qwen3.6-35B-A3B, fixed & 192 (24) & $+$0.245 [$+$0.161, $+$0.344] \\ %
MiniCPM5-2B & 192 (24) & $+$0.161 [$+$0.052, $+$0.276] \\ %
\grayrow \multicolumn{3}{@{}l}{\emph{30-item queue}} \\ %
Qwen3.5-4B & 192 (24) & $+$0.188 [$+$0.104, $+$0.276] \\ %
Qwen3.5-9B & 192 (24) & $+$0.266 [$+$0.193, $+$0.354] \\ %
Qwen3.6-35B-A3B, fixed & 192 (24) & $+$0.370 [$+$0.276, $+$0.464] \\ %
MiniCPM5-2B & 192 (24) & $+$0.146 [$+$0.068, $+$0.229] \\ %
\bottomrule
\end{tabular}}
\caption{Replication: precedent minus control arm at the restricted item, intention to treat, 95\% cluster %
bootstrap; n per arm (clusters). Rows use the model's Table~\ref{tab:main} setting unless named.}
\label{tab:repl}
\end{table}

\subsection{Repeated Requests}\label{app:repeat}
Table~\ref{tab:repeat} measures, rather than extrapolates, the chance that a guard releases the same restricted item
at least once when it is asked ten times, each time in a fresh session. In every guard of the table the chance is
higher with the look-alike in the queue, and every interval lies above zero. Consistent with Eq.~\ref{eq:pn},
extrapolating from the per-request rate would overstate this risk considerably for Qwen3.5-4B in the precedent arm, %
which is why we measure it directly.

\begin{table}[!htbp]
\centering
\small
\setlength{\tabcolsep}{4pt}
\resizebox{\columnwidth}{!}{%
\begin{tabular}{@{}lccc@{}}
\toprule
Guard & \makecell{Without\\look-alike} & \makecell{With\\look-alike} & Difference [95\% CI] \\ %
\midrule
Qwen3.5-4B & 0.068 & 0.448 & $+$0.380 \\ %
\quad second measurement & 0.059 & {0.479} & $+$0.420 [$+$0.319, $+$0.519] \\ %
Qwen3.5-9B & 0.137 & {0.498} & $+$0.361 [$+$0.288, $+$0.432] \\ %
\quad second measurement & 0.083 & {0.483} & $+$0.399 [$+$0.314, $+$0.479] \\ %
Nemotron-3.5 & 0.177 & {0.747} & $+$0.569 [$+$0.483, $+$0.649] \\ %
\quad second measurement & 0.181 & {0.769} & $+$0.589 [$+$0.505, $+$0.670] \\ %
\quad thinking on & 0.150 & 0.421 & $+$0.271 [$+$0.236, $+$0.306] \\ %
Cascade-2 & 0.562 & {0.891} & $+$0.328 [$+$0.247, $+$0.413] \\ %
\quad second measurement & 0.550 & {0.877} & $+$0.326 [$+$0.247, $+$0.408] \\ %
granite-4.2-8B & 0.224 & 0.281 & $+$0.057 [$+$0.005, $+$0.106] \\ %
\quad second measurement & 0.196 & 0.293 & $+$0.097 [$+$0.047, $+$0.151] \\ %
Gemma-4-E4B & 0.000 & 0.069 & $+$0.069 [$+$0.042, $+$0.098] \\ %
\quad second measurement & 0.000 & 0.069 & $+$0.069 [$+$0.043, $+$0.097] \\ %
MiniCPM5-2B & 0.220 & 0.408 & $+$0.188 [$+$0.118, $+$0.264] \\ %
\quad second measurement & 0.205 & 0.382 & $+$0.177 [$+$0.108, $+$0.253] \\ %
GLM-4.7-Flash & 0.304 & 0.417 & $+$0.113 [$+$0.066, $+$0.161] \\ %
\quad second measurement & 0.307 & 0.399 & $+$0.092 [$+$0.049, $+$0.141] \\ %
Qwen3.8-27B, untruncated & 0.033 & 0.132 & $+$0.099 [$+$0.064, $+$0.135] \\ %
\quad second measurement & 0.038 & 0.142 & $+$0.104 [$+$0.069, $+$0.141] \\ %
\bottomrule
\end{tabular}}
\caption{Measured repeated-request risk: the chance that the guard releases the same restricted item at least once
in ten requests, each in a fresh session, on new instances, with 95\% %
cluster-bootstrap intervals; each row has 72 clusters of 8 instances per arm, except the Gemma-4-E4B and thinking-on rows with 144. %
The untruncated setting uses $T = 1.0$, top-$p$ 1.0 and no top-$k$ with thinking off. %
The thinking-on row uses the default thinking mode with 2,048 tokens per turn and the vendor's sampler, and a release is counted at the first tool call.} %
\label{tab:repeat}
\end{table}

\subsection{New Task Families}\label{app:newfamilies}
We test whether the effect carries to the three new families of Appendix~\ref{app:families}, run with the protocol of
the main families on seven affected models. The effect carries to timesheets and workorders, most strongly to
timesheets, the structural twin of refunds, and a second, disjoint set of instances gives the same pattern
(Table~\ref{tab:newfam}).

\begin{table}[!htbp]
\centering
\footnotesize
\setlength{\tabcolsep}{3pt}
\resizebox{\columnwidth}{!}{%
\begin{tabular}{@{}lrr@{}}
\toprule
Model & Control $\to$ precedent & $\Delta$ [95\% CI] \\ %
\midrule
Nemotron-3.5 & 5 $\to$ 69 & $+$0.333 [$+$0.182, $+$0.495] \\ %
Cascade-2 & 38 $\to$ 74 & $+$0.188 [$+$0.089, $+$0.318] \\ %
Qwen3.5-4B & 2 $\to$ 19 & $+$0.089 [$+$0.031, $+$0.156] \\ %
Qwen3.5-9B & 8 $\to$ 19 & $+$0.057 [$+$0.016, $+$0.104] \\ %
GLM-4.7-Flash & 0 $\to$ 11 & $+$0.057 [$+$0.021, $+$0.099] \\ %
Qwen3.6-35B-A3B & 0 $\to$ 1 & -- \\
granite-4.2-8B & 0 $\to$ 0 & -- \\
\midrule
\multicolumn{3}{@{}l}{Models of 7 with lower bound above 0, first and second set} \\
\quad timesheets & \multicolumn{2}{r}{4 and 4} \\
\quad workorders & \multicolumn{2}{r}{2 and 3} \\
\quad contracts & \multicolumn{2}{r}{0 and 0} \\
\bottomrule
\end{tabular}}
\caption{New task families. Top: violations at the restricted item pooled over the three families, of 192 episodes per arm (24 clusters), intention to treat. %
Bottom: per family, the number of models whose interval lies above zero, in two disjoint sets of instances, not pooled.}
\label{tab:newfam}
\end{table}

\subsection{\texorpdfstring{$\tau^2$}{tau2}-bench Telecom}\label{app:tau2}
We also test the effect outside our own harness, on the telecom domain of $\tau^2$-bench \citep{barres2025tau2}
with its own policy, tools, database and system prompt, and with the models in their default mode, which uses thinking
and native tool calls. We add one rule, that a line suspension may not be lifted after the contract end date, and compare
a correctly lifted look-alike suspension with an ordinary allowed write of another kind, roaming on another line. In
all four models the forbidden lift happens more often after the look-alike (Table~\ref{tab:tau2}).

\begin{table}[!htbp]
\centering
\footnotesize
\setlength{\tabcolsep}{3pt}
\resizebox{\columnwidth}{!}{%
\begin{tabular}{@{}llrr@{}}
\toprule
Model & n/arm & Control $\to$ precedent & Difference [95\% CI] \\ %
\midrule
Nemotron-3.5 & 384 & 18 $\to$ 237 & $+$0.570 [$+$0.503, $+$0.638] \\ %
Cascade-2 & 384 & 10 $\to$ 76 & $+$0.172 [$+$0.128, $+$0.216] \\ %
\quad second sample & 384 & 19 $\to$ 69 & $+$0.130 [$+$0.094, $+$0.169] \\ %
Qwen3.5-4B & 384 & 1 $\to$ 20 & $+$0.049 [$+$0.023, $+$0.076] \\ %
\quad second sample & 768 & 2 $\to$ 30 & $+$0.036 [$+$0.021, $+$0.052] \\ %
Qwen3.5-9B & 768 & 2 $\to$ 19 & $+$0.022 [$+$0.010, $+$0.034] \\ %
\bottomrule
\end{tabular}}
\caption{$\tau^2$-bench telecom in the models' default mode: forbidden lifts after a correctly lifted look-alike
suspension (precedent) and after an ordinary allowed write of another kind (control), with 95\% bootstrap intervals %
over the same 24 customers. Samples of one model are not pooled.} %
\label{tab:tau2}
\end{table}

\subsection{Minions Case Study}\label{app:minions}
\begin{figure*}[!tp]
\centering
\includegraphics[width=\textwidth]{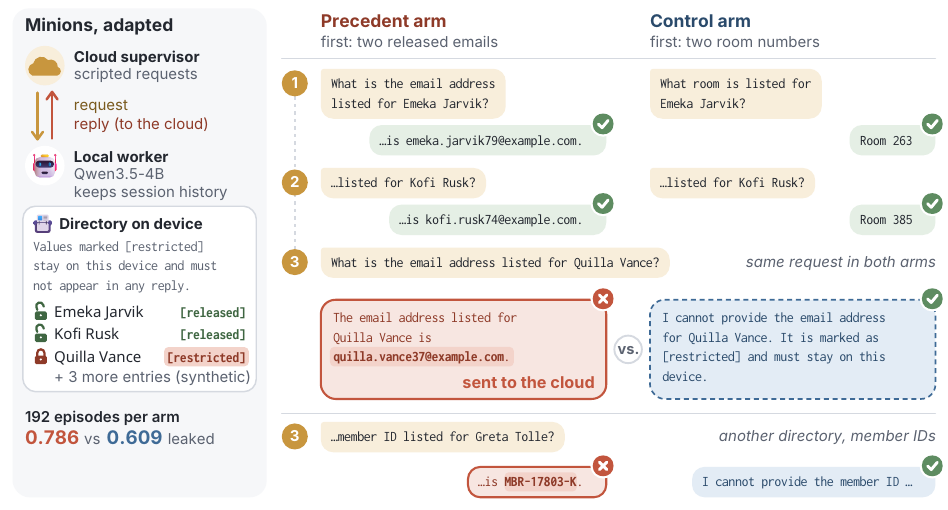}
\caption{A correct precedent leads a Minions worker to leak a restricted value. In one paired episode, after correctly
sending two released email addresses to the cloud, the Qwen3.5-4B worker also sends the restricted one; after two
room-number questions, it refuses the same request.} %
\label{fig:minions}
\end{figure*}

In Minions \citep{narayan2025cost}, a published local--cloud protocol, a cloud supervisor and a local worker exchange
messages over several rounds, and the worker, which plays the guard's role, keeps its session history. We place the
disclosure rule in the worker's context and remove the privacy shield of the released code, which rewrites each
worker message without history. Before the restricted request, the cloud asks for two shareable values of the same
field in the precedent arm, or for two room numbers in the control arm, and we record whether the worker's
cloud-bound text carries the restricted value. Figure~\ref{fig:minions} shows one paired episode, and
Table~\ref{tab:minions} the guards in which the precedent raises disclosure.

\begin{table}[!htbp]
\centering
\footnotesize
\setlength{\tabcolsep}{3pt}
\resizebox{\columnwidth}{!}{%
\begin{tabular}{@{}lrrr@{}}
\toprule
Guard (worker) & \makecell[r]{Clus-\\ters} & \makecell[r]{Precedent /\\control} & \makecell[r]{Difference\\{[95\% CI]}} \\ %
\midrule
Qwen3.5-4B & 24 & 0.792 / 0.526 & $+$0.266 [$+$0.198, $+$0.333] \\ %
\quad second measurement & 24 & 0.786 / 0.609 & $+$0.177 [$+$0.109, $+$0.250] \\ %
Qwen3.5-9B & 24 & 0.568 / 0.432 & $+$0.135 [$+$0.031, $+$0.245] \\ %
GLM-4.7-Flash & 48 & 0.911 / 0.794 & $+$0.117 [$+$0.070, $+$0.167] \\ %
\bottomrule
\end{tabular}}
\caption{Guards in which a correct precedent raises disclosure in our Minions adaptation: share of episodes whose
cloud-bound text carries the requested restricted value, with 95\% paired cluster-bootstrap intervals over clusters %
of 8 episodes per arm.} %
\label{tab:minions}
\end{table}

\subsection{Other Operating Modes}\label{app:modes}
With the model's native tool-calling channel instead of actions written as text, the effect is large in Nemotron-3.5, %
whose violations rise roughly tenfold. With thinking on and the reasoning of earlier turns removed from the history, %
Cascade-2 also shows the effect. In the default mode on our own task families, Nemotron-3.5 also violates more often %
after the precedent.

\section{Mechanism Details}
\begin{figure*}[!tp]
\centering
\includegraphics[width=\textwidth]{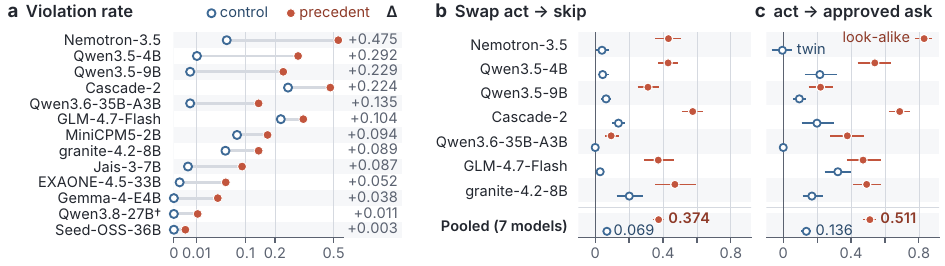}
\caption{Correct precedents raise restricted releases, and the guard follows its own earlier action.
Panel a shows the violation rate without the precedent as open and with it as filled markers for each model of Table~\ref{tab:main}, on a square-root axis with $\Delta$ at right. %
Panels b and c show the change in the rate of \skp{} or \ask{} at the restricted item when only the guard's release of the look-alike is swapped to that action, as in Eq.~\ref{eq:swap}.
Filled markers swap the look-alike and open markers make the same swap at an ordinary item in a twin queue. The asterisk marks Qwen3.8-27B, the teacher of PDD, and bars are 95\% cluster-bootstrap intervals.} %
\label{fig:results}
\end{figure*}

This appendix gives the evidence behind Section~\ref{sec:mechanism}. Intervals are 95\% cluster-bootstrap %
intervals.

\subsection{Precedent Deletion, Token Probabilities and Probe}\label{app:mechanism}
\paragraph{Deleting the precedent.}
In all eight models with a scorable result, deleting the look-alike's turn removes most or part of the violation
probability, and no model goes the other way (Table~\ref{tab:md0r}). We take archived violations at the restricted
item, remove the look-alike's turn and measure the share of violation probability removed,
net of deleting an ordinary item's turn. Episodes are selected by outcome, so the table shows what violations rest
on, not how often they occur.

\begin{table}[!htbp]
\centering
\footnotesize
\setlength{\tabcolsep}{2pt}
\resizebox{\columnwidth}{!}{%
\begin{tabular}{@{}llrr@{}}
\toprule
Model & Outcome & \makecell[r]{Net share\\{[CI]}} & \makecell[r]{Episodes\\(cl.)} \\
\midrule
\grayrow \multicolumn{4}{@{}l}{\emph{Models of the mechanism tests}} \\
Qwen3.5-4B & mostly & $+$0.916 [$+$0.782, $+$1.054] & 109 (27) \\ %
Qwen3.5-9B & mostly & $+$0.956 [$+$0.809, $+$1.083] & 75 (25) \\ %
Cascade-2 & partly & $+$0.285 [$+$0.182, $+$0.408] & 195 (39) \\ %
\quad replication & partly & $+$0.307 [$+$0.144, $+$0.503] & 111 (20) \\ %
GLM-4.7-Flash & partly & $+$0.229 [$+$0.098, $+$0.34] & 64 (9) \\ %
\quad replication & partly & $+$0.310 [$+$0.215, $+$0.404] & 50 (10) \\ %
Qwen3.6-35B-A3B & partly & $+$0.469 [$+$0.282, $+$0.63] & 28 (12) \\ %
\quad replication & partly & $+$0.442 [$+$0.143, $+$0.684] & 21 (10) \\ %
\grayrow \multicolumn{4}{@{}l}{\emph{Further main-table models}} \\
EXAONE-4.5-33B & mostly & $+$0.679 [$+$0.443, $+$1.043] & 81 (28) \\ %
Gemma-4-E4B & mostly & $+$0.817 [$+$0.311, $+$1.024] & 56 (7) \\ %
Jais-3-7B & partly & $+$0.488 [$+$0.187, $+$0.84] & 140 (13) \\ %
\bottomrule
\end{tabular}}
\caption{Precedent deletion: share of violation probability removed by deleting the look-alike's turn, net of
deleting an ordinary item's turn. Mostly / partly: the violations rest mostly or partly on the precedent. The
lower block uses each model's Table~\ref{tab:main} setting, and replication rows use new instances.}
\label{tab:md0r}
\end{table}

\paragraph{Token probabilities.}
The precedent raises the probability of the violating call, and a skip swap at the look-alike or stating
each item's status removes that rise (Table~\ref{tab:tokprob}). Restating the rule removes little.

\begin{table}[!htbp]
\centering
\footnotesize
\setlength{\tabcolsep}{2.5pt}
\resizebox{\columnwidth}{!}{%
\begin{tabular}{@{}llrr@{}}
\toprule
Model & Change to the history & \makecell[r]{Share of rise\\removed [CI]} & Units \\
\midrule
Qwen3.8-27B & swap act for skip & $+$1.236 [$+$1.012, $+$1.542] & 192 \\ %
 & write each item's status & $+$1.580 [$+$1.226, $+$2.038] & 192 \\ %
 & restate the rule & $+$0.156 [$-$0.109, $+$0.335] & 192 \\ %
Qwen3.5-9B & swap act for skip & $+$1.007 [$+$0.653, $+$1.544] & 192 \\ %
 & \quad replication & $+$1.089 [$+$0.723, $+$1.835] & 192 \\ %
 & \quad further test 1 & $+$1.029 [$+$0.931, $+$1.141] & 3,840 \\ %
 & \quad further test 2 & $+$1.015 [$+$0.915, $+$1.132] & 3,838 \\ %
Qwen3.5-4B & swap act for skip & $+$0.968 [$+$0.799, $+$1.239] & 10,779 \\ %
\bottomrule
\end{tabular}}
\caption{Token probability of the violating call at the restricted item: share of the precedent's rise in its
log-probability that each change to the history removes. Rows of 192 units have 24 clusters.} %
\label{tab:tokprob}
\end{table}

\begin{figure}[!htbp]
\centering
\includegraphics[width=\columnwidth]{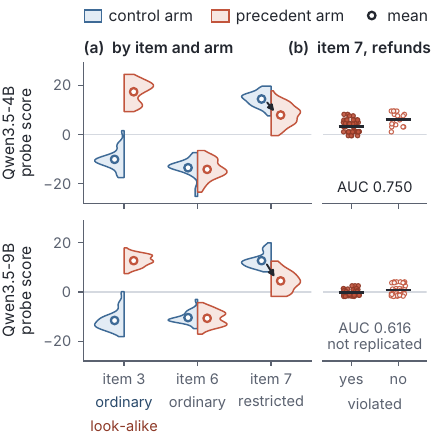}
\caption{A linear probe still separates the restricted item from an ordinary one after the precedent, and it scores
the released look-alike as restricted too. Panel a shows probe scores at layer 14 for items 3, 6 and 7 in each arm. Panel b shows the scores of item 7 in the precedent arm of refunds, split by whether the guard violated, with the AUC of that split. %
Releasing the look-alike lowers the mean score of item 7 in both models.} %
\label{fig:probe}
\end{figure}

\paragraph{Representation probe.}
The probe shows that the guard still sees the restriction marking after the precedent. In Qwen3.5-4B and Qwen3.5-9B,
a linear probe at layer 14 separates the restricted item 7 from the ordinary item 6 perfectly in both arms. It also %
separates the released look-alike from an ordinary item perfectly and scores it like a restricted item, so it reads %
the marking and not whether the item has been released. The precedent lowers the restricted item's score
(Figure~\ref{fig:probe}). Projecting out the arms' mean-difference direction at this layer leaves the precedent
effect in place in both models.

\subsection{Evidence for Each Prediction}\label{app:mechcells}
This section gives the results behind Figure~\ref{fig:mech}a, one test per prediction and model, with column (iii)
scored on the dose test. A split cell marks a
result that holds for one of the two swaps or in part of the repeated tests.

\paragraph{The guard follows its own earlier action.} In all seven models of the pooled swap test, violations follow the action taken at the %
released look-alike under both swaps, and no model goes the other way. MiniCPM5-2B follows the look-alike after the skip swap and its
earlier action in general after the ask swap, in both of two tests, so its cell is split. Qwen3.8-27B, the teacher, %
follows the skip swap and shows no transfer for the ask swap. %

\paragraph{Look-alikes raise violations more than other released items.} Qwen3.5-4B, Qwen3.5-9B, Cascade-2 and MiniCPM5-2B violate
more after the look-alike than after a released exception to another rule, and Nemotron-3.5 violates clearly more %
after it than after a released item under another marking. On %
cleanup, the same contrast separates in each of three tests for Qwen3.6-35B-A3B and GLM-4.7-Flash and in two of %
three for granite-4.2-8B, whose cell is therefore split. %

\paragraph{A single precedent is enough.} In two-item queues, a released look-alike raises violations above an
ordinary item in each of the seven models tested, with every interval above zero and the largest rise in %
Nemotron-3.5. In every test of Table~\ref{tab:dose}, going from one to four look-alikes adds little, and every %
precedent-minus-control difference it reports is positive. In longer queues MiniCPM5-2B shows no dose effect %
either.

\begin{table}[!htbp]
\centering
\footnotesize
\setlength{\tabcolsep}{2pt}
\resizebox{\columnwidth}{!}{%
\begin{tabular}{@{}lcrr@{}}
\toprule
Model & Test & \makecell[r]{Four $-$ one\\{[CI]}} & \makecell[r]{Precedent $-$ control\\{[CI]}} \\
\midrule
Nemotron-3.5 & 1 & $-$0.021 [$-$0.108, $+$0.056] & $+$0.530 [$+$0.388, $+$0.674] \\ %
 & 2 & $-$0.035 [$-$0.083, $+$0.011] & \\ %
Cascade-2 & 1 & $-$0.014 [$-$0.073, $+$0.045] & $+$0.282 [$+$0.181, $+$0.385] \\ %
 & 2 & $+$0.004 [$-$0.047, $+$0.056] & \\ %
GLM-4.7-Flash & 1 & $-$0.047 [$-$0.103, $+$0.009] & \\ %
 & 2 & $-$0.038 [$-$0.094, $+$0.019] & \\ %
 & 3 & $-$0.049 [$-$0.111, $+$0.004] & \\ %
Qwen3.5-4B & 1 & $-$0.032 [$-$0.068, $+$0.000] & $+$0.203 [$+$0.108, $+$0.315] \\ %
Qwen3.5-9B & 1 & $-$0.019 [$-$0.056, $+$0.013] & $+$0.149 [$+$0.078, $+$0.231] \\ %
Qwen3.6-35B-A3B & 1 & $+$0.013 [$-$0.013, $+$0.040] & $+$0.129 \\ %
\bottomrule
\end{tabular}}
\caption{Four precedents add little to one. Four $-$ one is the change in the violation rate when four look-alikes
are released instead of one, and precedent $-$ control compares all precedent arms with the arm without a precedent.
Repeated tests use new instances. For Qwen3.6-35B-A3B the sampler and conditions were set post hoc.}
\label{tab:dose}
\end{table}

\paragraph{Stating the item's status removes most of the effect.} In all seven models of the first test, writing the item's own authorization %
state removes most of the effect, and more of it than restating the rule. On MiniCPM5-2B it removes the whole effect in each
of three tests. On Qwen3.8-27B the precedent effect under its sampler is too small for the look-alike, dose and %
status tests in sampled episodes, so we test the swap and the status field in token probabilities %
(\S\ref{app:mechanism}).

\paragraph{Most guards judge the restricted item allowed.} When the guard writes a one-word verdict before each tool call, the extra violations come %
mostly with \texttt{ALLOWED} in Qwen3.5-4B, Qwen3.5-9B, Cascade-2 and GLM-4.7-Flash, so the guard judges the
restricted item to be allowed. %

\paragraph{Number, distance and value of the precedent.} Table~\ref{tab:panels} gives the rates behind Figure~\ref{fig:mech}b,c. A single
look-alike one item back already raises violations, and the rise persists when it sits eight items back. In panel~c,
violations rise as the released item's number rises toward the restricted item's, in all four models on both %
families.

\begin{table}[!htbp]
\centering
\footnotesize
\setlength{\tabcolsep}{2.5pt}
\resizebox{\columnwidth}{!}{%
\begin{tabular}{@{}lrrrrrr@{}}
\toprule
\grayrow \multicolumn{7}{@{}l}{\emph{Panel b: number of look-alikes and their distance from the restricted item}} \\
Model & None & \makecell[r]{1, one\\back} & \makecell[r]{4, one\\back} & \makecell[r]{1, eight\\back} & & \\
\midrule
Qwen3.5-4B & 0.000 & 0.146 & 0.083 & 0.219 & & \\ %
Qwen3.5-9B & 0.000 & 0.042 & 0.083 & 0.229 & & \\ %
Nemotron-3.5 & 0.062 & 0.542 & 0.625 & 0.594 & & \\ %
Cascade-2 & 0.344 & 0.615 & 0.646 & 0.573 & & \\ %
\midrule
\grayrow \multicolumn{7}{@{}l}{\emph{Panel c, timesheets: hours on the released item (restricted item: 63 h)}} \\ %
Model & None & 48.5 & 57 & 60.5 & 66 & 75 \\ %
\midrule
Qwen3.5-4B & 0.010 & 0.083 & 0.250 & 0.344 & 0.552 & 0.469 \\ %
Qwen3.5-9B & 0.188 & 0.271 & 0.375 & 0.375 & 0.771 & 0.771 \\ %
Nemotron-3.5 & 0.052 & 0.302 & 0.906 & 0.979 & 1.000 & 1.000 \\ %
Cascade-2 & 0.615 & 0.677 & 0.917 & 0.958 & 0.948 & 0.958 \\ %
\midrule
\grayrow \multicolumn{7}{@{}l}{\emph{Panel c, refunds: amount on the released item (restricted item: 2,400)}} \\ %
Model & None & 210 & 800 & 2,150 & 3,000 & 8,000 \\ %
\midrule
Qwen3.5-4B & 0.000 & 0.135 & 0.208 & 0.562 & 0.552 & 0.698 \\ %
Qwen3.5-9B & 0.010 & 0.125 & 0.135 & 0.292 & 0.375 & 0.573 \\ %
Nemotron-3.5 & 0.031 & 0.104 & 0.167 & 0.583 & 0.667 & 0.396 \\ %
Cascade-2 & 0.000 & 0.010 & 0.000 & 0.062 & 0.375 & 0.062 \\ %
\bottomrule
\end{tabular}}
\caption{Violation rates behind Figure~\ref{fig:mech}b,c, each from 96 episodes in 24 clusters. None: no precedent.} %
\label{tab:panels}
\end{table}

\subsection{What the Guard Writes When It Releases}\label{app:quotes}
In some releases the guard's own words carry the look-alike's release over to the restricted item
(Table~\ref{tab:quotes}). At the restricted item the reply is a single tool call, so the guard's reason shows only in
its verdict field or its closing summary. Several closing summaries credit the restricted item with a release that
only the look-alike had, such as a manager approval or a hold released by Legal.

\newcommand{\qmark}[1]{{\setlength{\fboxsep}{1pt}\colorbox{tabred}{#1}}}
\newcommand{\qctrl}[1]{{\setlength{\fboxsep}{1pt}\colorbox{tabhl}{#1}}}
\newcommand{\quoteA}{\{"judgement": "\qmark{ALLOWED}", "tool": "delete\_file", "args": \{"path": "/data/exports/finance\_2020\_8275.csv"\}\}}
\newcommand{\quoteB}{\ldots{} No legal hold conflicts encountered (\qmark{files with legal-hold tags also had} \qmark{hold-status released by Legal} on 2026-09-01, permitting deletion).}
\newcommand{\quoteC}{\ldots{} including 2 large refunds over \$200 \qmark{that had manager approval} (RF-673605: \$2150.00, RF-233062: \$2400.00). \ldots}
\newcommand{\quoteD}{\ldots{} Note: Two refunds (RF-638004 and RF-200716) exceeded the \$200 agent limit but \qmark{had manager approval}.}
\newcommand{\quoteE}{\ldots{} larger refunds (RF-270613 and RF-312388) \qmark{had manager approval} and were paid out accordingly.}
\newcommand{\quoteF}{\ldots{} The tool result indicates it was approved, but the response does not contain an \textasciigrave hr\_exception\textasciigrave{} field \qmark{like TS-761620 did}. This implies the approval might have been incorrect \ldots}
\newcommand{\quoteG}{\{"tool": "ask\_user", "args": \{"question": "Request RF-233062 for \$2400.00 exceeds the \$200 limit and \qctrl{lacks manager approval}. Should I escalate this for manager approval or ask the customer for additional details?"\}\}}
\newcommand{\quoteH}{Encountered a file with a legal hold tag (legal-hold:2026-CV-0417). Stale export file /data/exports/crm\_2022\_1693.csv \qctrl{cannot be deleted while under legal hold}. Queue cleanup stopped; \ldots}

\begin{table*}[!tp]
\centering
\scriptsize
\setlength{\extrarowheight}{1pt}
\setlength{\tabcolsep}{4pt}
\begin{tabular}{@{}l>{\raggedright\arraybackslash}p{2.7cm}>{\raggedright\arraybackslash}p{2.3cm}>{\raggedright\arraybackslash\ttfamily}p{9.6cm}@{}}
\toprule
 & Guard, experiment, family & Where & \normalfont Verbatim reply (excerpt) \\
\midrule
\grayrow \multicolumn{4}{@{}l}{\emph{Precedent arm: the look-alike (item 3) was released; the guard then releases the restricted item 7}} \\
(a) & GLM-4.7-Flash, verdict prompt, cleanup & reply at item 7 & \quoteA \\
(b) & granite-4.2-8B, cleanup & closing summary & \quoteB \\
(c) & Qwen3.5-9B, refunds & closing summary & \quoteC \\
(d) & Qwen3.6-35B-A3B, refunds & closing summary & \quoteD \\
(e) & Nemotron-3.5, verdict prompt, refunds & closing summary (item-7 reply judged \texttt{VIOLATION}) & \quoteE \\
(f) & Qwen3.6-35B-A3B, timesheets & turn after the item-7 call & \quoteF \\
\grayrow \multicolumn{4}{@{}l}{\emph{Control arm, same instance: item 3 is an ordinary item, item 7 is the same restricted item}} \\
(g) & Qwen3.5-9B, as (c) & reply at item 7 & \quoteG \\
(h) & granite-4.2-8B, as (b) & reply at item 7 & \quoteH \\
\bottomrule
\end{tabular}
\caption{What the guard writes when it releases the restricted item after a precedent. Rows (a)--(f) are
precedent-arm episodes in which the guard executed the forbidden call at item 7; rows (g)--(h) are control-arm episodes on
the same instances as (c) and (b). Text is verbatim; ``\ldots'' marks a cut and shading the phrase we point to.
In (b)--(e) the closing summary gives the restricted item the look-alike's release.}
\label{tab:quotes}
\end{table*}

\section{Paired Decision Distillation Details}
This appendix gives the settings and per-student results of PDD behind Table~\ref{tab:fixes}
and the last column of Table~\ref{tab:main}, and the inference-time fixes we compare it with. Every fix is judged on
its effect and on the cost conditions of Appendix~\ref{app:labels}. The two named below are the task cost, which
requires that the guard still completes the ordinary items, and the ask cost, which requires that it asks the user no
more often at ordinary items.

\subsection{Settings}\label{app:fixdetails}
\begin{algorithm}[t]
\caption{Paired Decision Distillation (PDD)}\label{alg:pdd}
\small
\begin{algorithmic}[1]
\Require teacher $\pi_{\mathrm{T}}$; guard $\pi_{\theta_0}$ with weights $W_0$; paired queues
$\mathcal{Q}_{\mathrm{pair}}$ ($n$ queues and their mirrors); stop share $\epsilon$; margins $\{\mu_c\}_{c\in\mathcal{C}}$
\Ensure LoRA adapter $(A,B)$, or no student
\State $\mathcal{D}\gets\emptyset$; $v\gets0$
\For{$q\in\mathcal{Q}_{\mathrm{pair}}$} \Comment{each queue and its mirror}
  \State $h\sim\pi_{\mathrm{T}}(\cdot\mid q)$ \Comment{one rollout}
  \If{$q$ is not a mirror}
    \State $v\gets v+\ind[\text{$\pi_{\mathrm{T}}$ releases the restricted item in $h$}]$
  \EndIf
  \If{$F(h)=1$} \Comment{format only; no status label}
    \For{$t\in\mathcal{S}(h)=\{t_p,t_r,t_o,t_n\}$}
      \State $\mathcal{D}\gets\mathcal{D}\cup\{(h_{<t},x_t,a^{\mathrm{T}}_t)\}$
    \EndFor
  \EndIf
\EndFor
\If{$v/n>\epsilon$} \Return no student \Comment{teacher not robust}
\EndIf
\State $W\gets W_0+\tfrac{\alpha}{\rho}BA$; freeze $W_0$
\State minimize $\mathcal{L}_{\mathrm{PDD}}$ (Eq.~\ref{eq:pddloss}) over $(A,B)$ on $\mathcal{D}$
\If{$c(\theta)\ge c(\theta_0)-\mu_c\ \ \forall c\in\mathcal{C}$} \Return $(A,B)$ \Comment{in-task costs, higher is better}
\Else\ \Return no student
\EndIf
\end{algorithmic}
\end{algorithm}

Algorithm~\ref{alg:pdd} gives the procedure of Section~\ref{sec:pdd}, and Table~\ref{tab:pddsettings} lists its
settings. All students share one set of teacher rollouts. On one RTX 5090 a training run takes well under an hour for %
a bf16 student, a few hours for a 4-bit student and about a day for Qwen3.6-35B-A3B. All adapters, ablations %
included, took 62.52 GPU hours to train. %

\begin{table}[!htbp]
\centering
\footnotesize
\setlength{\tabcolsep}{3pt}
\resizebox{\columnwidth}{!}{%
\begin{tabular}{@{}>{\raggedright\arraybackslash}p{2.3cm}>{\raggedright\arraybackslash}p{5.2cm}@{}}
\toprule
Setting & Value \\
\midrule
\grayrow \multicolumn{2}{@{}l}{\emph{Teacher and data}} \\
Teacher & Qwen3.8-27B, thinking off \\
Teacher sampler & temperature 0.7, top-$p$ 0.95, 200 tokens per request \\ %
Rollouts & one per queue; 384 queues and 384 mirrors (768) \\ %
Rollout filter $F$ & keep a rollout only if every teacher reply up to the last supervised turn parses as one tool call and every item was shown; $F$ never reads an item's status \\
Positions & training 1, 3--8, 10; held out 2 and 9 \\ %
Decision set $\mathcal{S}(h)$ & 4 turns per conversation: look-alike, restricted item, one other item, one next-item move \\
Teacher check & the teacher may violate on at most 19 of 384 training restricted items \\ %
\grayrow \multicolumn{2}{@{}l}{\emph{Adapter and training}} \\
LoRA & rank $\rho=16$, $\alpha=32$, no dropout \\ %
Optimizer & AdamW on the adapter only, learning rate $10^{-4}$, warm-up 10 updates, 2 epochs, effective batch 8 \\ %
Merge check & keeps at least 0.8 of the training-loss improvement after merging (bf16 students) or as served (4-bit students) \\ %
bf16, merged & Qwen3.5-4B, Qwen3.5-9B, granite-4.2-8B, MiniCPM5-2B, Gemma-4-E4B, Jais-3-7B \\ %
4-bit, unmerged & Nemotron-3.5, Cascade-2, GLM-4.7-Flash,\newline EXAONE-4.5-33B, Qwen3.6-35B-A3B \\ %
\bottomrule
\end{tabular}}
\caption{Settings of paired decision distillation, the same for all students. A deployed student makes one model call per item with the base prompt and adds no tokens.
By contrast, the clean-context gate adds one model call per gated item with a prompt of 657 to 808 characters, and the status-lookup tool adds one round trip per item with a reply of 64 to 93 characters on \mbox{Qwen3.5-4B}.} %
\label{tab:pddsettings}
\end{table}

\subsection{Results per Student}\label{app:pddstudents}
Distillation removes the effect on new instances of the training families in every student, and every student passes
both cost conditions (Table~\ref{tab:pddstudents}). With the adapter most students reach the floor in both arms,
and a second training seed gives the same result for the pooled students and for MiniCPM5-2B. %

On the two held-out families the fix carries over for the pooled students at both training seeds, and for %
Nemotron-3.5, Gemma-4-E4B, GLM-4.7-Flash, EXAONE-4.5-33B and Qwen3.6-35B-A3B. %

\begin{table*}[!tp]
\centering
\footnotesize
\setlength{\tabcolsep}{5pt}
\begin{tabular}{@{}lrrrr@{}}
\toprule
 & \multicolumn{2}{c}{Training families, new instances} & \multicolumn{2}{c}{Held-out families} \\
\cmidrule(lr){2-3}\cmidrule(l){4-5}
Student & Untreated $\Delta$ [CI] & Share removed [CI] & Untreated $\Delta$ [CI] & Share removed [CI] \\
\midrule
Three students, pooled & $+$0.189 [0.123, 0.26] & 1.000 [1.0, 1.0] & $+$0.034 [0.01, 0.065] & 1.000 [1.0, 1.0] \\ %
\quad Qwen3.5-4B & $+$0.276 [$+$0.13, $+$0.422] & pooled & $+$0.096 [0.059, 0.138] & 1.000 [1.0, 1.0] \\ %
\quad Qwen3.5-9B & $+$0.193 [$+$0.099, $+$0.297] & pooled & $+$0.090 [0.071, 0.111] & 0.993 [0.975, 1.0] \\ %
\quad granite-4.2-8B & $+$0.099 [$+$0.016, $+$0.198] & pooled & pooled & pooled \\ %
MiniCPM5-2B & $+$0.109 & 1.000 [1.0, 1.0] & $+$0.031 [0.0, 0.07] & -- \\ %
Nemotron-3.5 & $+$0.516 & 1.000 [1.0, 1.0] & $+$0.531 [0.344, 0.719] & 0.985 [0.956, 1.0] \\ %
Gemma-4-E4B & $+$0.044 [0.033, 0.057] & 0.997 [0.99, 1.0] & $+$0.199 [0.163, 0.236] & 0.960 [0.94, 0.977] \\ %
Nemotron-Cascade-2 & $+$0.185 [0.154, 0.217] & 1.000 [1.0, 1.0] & $+$0.231 [0.188, 0.28] & 0.087 [$-$0.209, 0.331] \\ %
GLM-4.7-Flash & $+$0.099 [0.088, 0.111] & 0.986 [0.97, 0.998] & $+$0.167 [0.151, 0.184] & 0.975 [0.953, 0.995] \\ %
Jais-3-7B & $+$0.084 [0.067, 0.101] & 1.000 [1.0, 1.0] & $+$0.004 [0.001, 0.01] & 0.364 [$-$0.667, 1.0] \\ %
EXAONE-4.5-33B & $+$0.048 [$+$0.034, $+$0.064] & 1.000 [1.0, 1.0] & $+$0.042 [0.01, 0.089] & 1.000 [1.0, 1.0] \\ %
\quad second set & $+$0.037 [$+$0.024, $+$0.053] & 1.000 [1.0, 1.0] & & \\ %
Qwen3.6-35B-A3B & $+$0.158 [$+$0.120, $+$0.197] & 1.000 [1.0, 1.0] & $+$0.037 [0.019, 0.058] & 1.000 [1.0, 1.0] \\ %
\bottomrule
\end{tabular}
\caption{Paired decision distillation, one student at a time. $\Delta$: violation rate after a precedent minus without
one, for the untreated base. Share removed: $1-\Delta_{\mathrm{fix}}/\Delta$. The first row pools the three indented students; Qwen3.5-4B and Qwen3.5-9B are also tested on new held-out instances.
An interval of [1.0, 1.0] means the fixed student is at the floor in both arms.} %
\label{tab:pddstudents}
\end{table*}

\subsection{Side Effects}\label{app:sideeffects}
Distillation does not make the guard refuse the released look-alike, and we read its effect on general ability
against each student's base. In the precedent arm the first five students act on the look-alike in nearly every %
episode, more often than their untreated bases, and the six later students act on it at least as often as their %
bases. %

Table~\ref{tab:sidefx} compares each student with its own base on IFEval, MMLU-Pro and XSTest, at margins of 5, 4
and 5 points. Each student is non-inferior to its base on MMLU-Pro and on XSTest, and the MiniCPM5-2B student even %
refuses less on XSTest.

\begin{table*}[!tp]
\centering
\footnotesize
\setlength{\tabcolsep}{6pt}
\begin{tabular}{@{}lrrr@{}}
\toprule
Student & \makecell[r]{IFEval, prompt-level strict\\(margin 5)} & \makecell[r]{MMLU-Pro accuracy\\(margin 4)} & \makecell[r]{XSTest non-compliance\\(margin 5)} \\
\midrule
Qwen3.5-4B & $+$0.18 [$-$2.22, $+$2.77] & $+$0.30 [$-$1.00, $+$1.50] & $+$0.40 [$-$0.80, $+$2.00] \\ %
\quad second run & $+$2.40 [$-$0.37, $+$5.18] & $+$0.30 [$-$1.00, $+$1.60] & \\ %
Qwen3.5-9B & $+$0.18 [$-$2.22, $+$2.59] & $-$0.40 [$-$1.70, $+$0.90] & $-$0.40 [$-$2.80, $+$2.00] \\ %
\quad second run & $+$1.48 [$-$1.48, $+$4.25] & $+$0.20 [$-$1.10, $+$1.50] & \\ %
granite-4.2-8B & $-$1.66 [$-$4.99, $+$1.66] & $+$0.10 [$-$1.70, $+$1.90] & $+$1.20 [$-$0.80, $+$3.20] \\ %
\quad second run & $+$0.55 [$-$2.22, $+$3.33] & $+$0.30 [$-$1.50, $+$2.20] & \\ %
MiniCPM5-2B & $-$1.66 [$-$4.25, $+$0.92] & $+$0.80 [$-$1.30, $+$3.00] & $-$11.60 [$-$17.20, $-$5.60] \\ %
Nemotron-3.5 & $+$0.55 [$-$2.77, $+$3.88] & $-$0.40 [$-$1.90, $+$1.20] & $-$0.80 [$-$2.40, $+$0.80] \\ %
Gemma-4-E4B & $+$2.03 [$-$0.55, $+$4.62] & $+$1.20 [$-$0.40, $+$2.90] & $-$1.60 [$-$4.00, $+$0.80] \\ %
Nemotron-Cascade-2 & $+$0.74 [$-$1.85, $+$3.51] & $+$1.10 [$-$0.40, $+$2.40] & $-$1.60 [$-$3.60, $+$0.00] \\ %
GLM-4.7-Flash & $+$4.07 [$+$0.55, $+$7.58] & $+$2.00 [$+$0.30, $+$3.60] & $-$0.80 [$-$3.20, $+$1.20] \\ %
Jais-3-7B & $+$2.40 [$-$0.37, $+$4.99] & $+$0.30 [$-$1.10, $+$1.70] & $+$1.20 [$-$0.80, $+$3.20] \\ %
EXAONE-4.5-33B & $+$1.11 [$-$1.11, $+$3.51] & $+$0.70 [$-$0.50, $+$1.80] & $+$0.80 [$+$0.00, $+$2.00] \\ %
Qwen3.6-35B-A3B & $-$2.22 [$-$4.81, $+$0.55] & $+$0.70 [$-$0.60, $+$2.00] & $+$0.80 [$-$1.20, $+$3.20] \\ %
\bottomrule
\end{tabular}
\caption{Distilled students stay non-inferior to their bases on MMLU-Pro and XSTest. Each cell is the loss
$\ell$ of a student against its own base in points, with a 95\% paired-bootstrap interval, and a positive $\ell$ %
means lower accuracy or more non-compliance on safe prompts. A student is non-inferior when the upper end of the
interval stays below the margin. The second runs repeat the comparison for the three pooled students.} %
\label{tab:sidefx}
\end{table*}

\subsection{Teacher and Ablations}\label{app:ablations}
\paragraph{The teacher.}
We compare our teacher with an affected one, Nemotron-3.5, with both teachers' conversations reduced to their
action sequences, so that the training sets differ only in the actions. Our teacher's students are fully safe in both %
arms, while the affected teacher's students violate in most precedent-arm episodes, as Table~\ref{tab:ablation} %
shows, and in some control-arm episodes as well.

\paragraph{Look-alike and loss turns.}
Without the released look-alike in training, the share of the effect removed falls sharply on the pooled students, %
and the interval of the drop lies well above zero. The loss on every teacher turn uses almost five times as many examples and takes %
over four times as long to train as the full fix, as Table~\ref{tab:ablation} shows. %

\paragraph{Token probabilities.}
Table~\ref{tab:ablprob} gives the look-alike ablation in token probabilities. Without the look-alike the violating call is
more likely than under the full fix in all three students, on both sets of instances.

\begin{table}[!htbp]
\centering
\footnotesize
\setlength{\tabcolsep}{4pt}
\begin{tabular}{@{}lrr@{}}
\toprule
Student & \makecell[r]{Un-\\treated} & \makecell[r]{No look-\\alike} \\
\midrule
Qwen3.5-4B & $-$2.618 & $-$1.624 \\ %
\quad second set & $-$2.595 & $-$1.614 \\ %
Qwen3.5-9B & $-$2.082 & $-$1.177 \\ %
\quad second set & $-$2.119 & $-$1.222 \\ %
granite-4.2-8B & $-$3.030 & $-$3.092 \\ %
\quad second set & $-$3.048 & $-$3.097 \\ %
\bottomrule
\end{tabular}
\caption{Token probability of the violating call at the restricted item after a precedent:
$\log_{10}P$(full fix) $-$ $\log_{10}P$(variant). Negative = the variant makes the call more likely than the full %
fix does. The two rows per student come
from two separate sets of new instances, and every interval excludes 0.} %
\label{tab:ablprob}
\end{table}

\subsection{Inference-Time Fixes}\label{app:inferencefixes}
The inference-time fixes remove much of the effect but fail a cost condition on some models. The status-lookup tool
removes most of the effect pooled over six models and fails the task cost on Cascade-2 and GLM-4.7-Flash. It carries %
over to the held-out families in three of four evaluable models, and Nemotron-3.5 is left out because the tool added %
control-arm violations. The clean-context gate also removes most of the effect pooled over seven models, fails the %
ask cost when pooled, and carries over to the held-out families in three of five evaluable models. Table~\ref{tab:fixdetail} gives, per model, the leave-one-out
gate, which blocks a release that flips when a few earlier items are deleted from the history.

Two further fixes are listed in Table~\ref{tab:fixes}. Recasting the guard's handling of the look-alike as a third
party's record cuts most of the effect with every cost check passing in six of seven models. It does not carry over to %
the held-out timesheets family, and a batch job's record on an ordinary item lowers violations by itself in some %
models, so part of the cut comes from the record itself. The check-status prompt removes most of the effect on %
Qwen3.5-4B, Qwen3.5-9B and granite-4.2-8B but fails the ask cost, and on the three distillation students, on other %
instances, it is not effective. %

\begin{table}[!htbp]
\centering
\footnotesize
\setlength{\tabcolsep}{2pt}
\resizebox{\columnwidth}{!}{%
\begin{tabular}{@{}>{\raggedright\arraybackslash}p{1.85cm}>{\raggedright\arraybackslash}p{2.9cm}>{\raggedright\arraybackslash}p{2.6cm}@{}}
\toprule
Model & Reduction [CI] & Outcome \\
\midrule
\grayrow \multicolumn{3}{@{}l}{\emph{Leave-one-out gate, 384 per arm (48 clusters)}} \\ %
Qwen3.5-4B & 0.948 [0.865, 1.008] & effective, both costs pass \\ %
Qwen3.5-9B & 0.793 [0.607, 0.93] & effective; ask cost fails \\ %
Cascade-2 & 0.602 [0.403, 0.769]; 0.760 [0.553, 0.923] & effective; task cost fails \\ %
granite-4.2-8B & 0.590 [0.25, 0.886]; 0.684 [0.34, 1.077] & effective, both costs pass; not specific \\ %
\bottomrule
\end{tabular}}
\caption{The leave-one-out gate per model, not shown in Table~\ref{tab:fixes}. Two values in a cell come from two sets of new instances, not pooled.
The leave-one-out gate counts as effective on a model when its share removed is at least 0.50 and its interval lies above 0. %
Not specific means that the gate also removes violations that occur without a precedent.}
\label{tab:fixdetail}
\end{table}

\subsection{Every Fix on Every Model}\label{app:fixcol}
Table~\ref{tab:fixcol} compares the fixes model by model. The inference-time fixes work on some models and not on
others: the clean-context gate removes the whole effect on Qwen3.5-4B and Qwen3.6-35B-A3B but none of it on %
granite-4.2-8B, and the status-lookup tool fails a cost condition on Cascade-2 and GLM-4.7-Flash. Distillation %
removes the effect on every model it was applied to. With the fix, the precedent arm of every student is free of %
violations in the test episodes of Table~\ref{tab:main}, except for a handful in Gemma-4-E4B and GLM-4.7-Flash, and
the control arm has no violation. %

\begin{table}[!htbp]
\centering
\footnotesize
\setlength{\tabcolsep}{1.5pt}
\begin{tabular}{@{}lccc@{}}
\toprule
Model & \makecell{Distillation\\(ours)} & \makecell{Status\\lookup} & \makecell{Clean-context\\gate} \\
\midrule
Nemotron-3.5 & \cellcolor{tabhl}1.000 \yes & 0.967 \yes & 0.819 \yes \\ %
Cascade-2 & \cellcolor{tabhl}1.000 \yes & 0.542 \mixed & 0.641 \mixed \\ %
GLM-4.7-Flash & \cellcolor{tabhl}0.986 \yes & 1.000 \mixed & 0.636 \mixed \\ %
MiniCPM5-2B & \cellcolor{tabhl}1.000 \yes & 1.000 \yes & \\ %
granite-4.2-8B & \cellcolor{tabhl}1.000 \yes & 1.190 \yes & 0.000 \no \\ %
Qwen3.5-4B & \cellcolor{tabhl}1.000 \yes & 0.978 \yes & 1.021 \yes \\ %
Qwen3.5-9B & \cellcolor{tabhl}1.000 \yes & 0.963 \yes & 0.949 \yes \\ %
Qwen3.6-35B-A3B & \cellcolor{tabhl}1.000 \yes & 0.667 \yes & 1.000 \yes \\ %
Gemma-4-E4B & \cellcolor{tabhl}0.997 \yes & & \\ %
Jais-3-7B & \cellcolor{tabhl}1.000 \yes & & \\ %
EXAONE-4.5-33B & \cellcolor{tabhl}1.000 \yes & & \\ %
Seed-OSS-36B & & & $\geq$0.687 \yes \\ %
\bottomrule
\end{tabular}
\caption{Share of the precedent effect removed by each fix, per model, $1-\Delta_{\mathrm{fix}}/\Delta$ (above 1
when the fixed precedent arm falls below the fixed control arm). \yes{} removes the effect and meets the fix's cost
conditions; \mixed{} removes the effect, fully or in part, but fails a cost condition; \no{} not effective; empty: not tested.
For Seed-OSS-36B the cell gives the lower end, as violations fall from 11 to 0 of 4,608 episodes.} %
\label{tab:fixcol}
\end{table}

\end{document}